\documentclass[lettersize,journal]{IEEEtran}
\usepackage{amsmath,amsfonts}
\usepackage{algorithmic}
\usepackage{algorithm}
\usepackage{array}
\usepackage[caption=false,font=normalsize,labelfont=sf,textfont=sf]{subfig}
\usepackage{textcomp}
\usepackage{stfloats}
\usepackage{url}
\usepackage{verbatim}
\usepackage{graphicx}
\usepackage{cite}
\usepackage[normalem]{ulem}
\usepackage{framed}
\usepackage{lipsum}
\usepackage{amsmath} 
\usepackage{booktabs}
\usepackage{multirow}
\usepackage{xcolor}
\usepackage{tabularx}
\usepackage{threeparttable}
\usepackage{makecell}
\begin{document}

\title{\textbf{Design of an In-Pipe Robot with Contact-Angle-Guided Kinematic Decoupling for Crosstalk-Suppressed Locomotion}}

\author{Min Yang, Yang Tian, Longchuang Li, Jun Ma, \textit{Senior Member, IEEE}, Shugen Ma, \textit{Fellow, IEEE}\\
	
\thanks{Min Yang, Jun Ma, and Shugen Ma are with the Robotics and Autonomous Systems Thrust, The Hong Kong University of Science and Technology (Guangzhou), Guangzhou 511453, China (e-mail: myang945@connect.hkust-gz.edu.cn; jun.ma@ust.hk; shugenma@hkust-gz.edu.cn).}
\thanks{Yang Tian is with the Department of Engineering, Shinshu University, Nagano 380-8553, Japan (e-mail: tian@shinshu-u.ac.jp).}
\thanks{Longchuan Li is with the College of Information Science and Technology, Beijing University of Chemical Technology, Beijing 100029, China (e-mail: longchuanli@buct.edu.cn).}

}



\maketitle

\begin{abstract}
In-pipe inspection robots must traverse confined pipeline networks with elbows and three-dimensional fittings, requiring both reliable axial traction and rapid rolling reorientation for posture correction. In compact V-shaped platforms, these functions often rely on shared contacts or indirect actuation, which introduces strong kinematic coupling and makes performance sensitive to geometry and friction variations. This paper presents a V-shaped in-pipe robot with a joint-axis-and-wheel-separation layout that provides two physically independent actuation channels, with all-wheel-drive propulsion and motorized rolling reorientation while using only two motors. To make the decoupling mechanism explicit and designable, we formulate an actuation transmission matrix and identify the spherical-wheel contact angle as the key geometric variable governing the dominant roll-to-propulsion leakage and roll-channel efficiency. A geometric transmission analysis maps mounting parameters to the contact angle, leakage, and efficiency, yielding a structural guideline for suppressing crosstalk by driving the contact angle toward zero. A static stability model further provides a stability-domain map for selecting torsion-spring stiffness under friction uncertainty to ensure vertical-pipe stability with a margin. Experiments validate the decoupling effect, where during high-dynamic rolling in a vertical pipe, the propulsion torque remains nearly invariant. On a multi-material testbed including out-of-plane double elbows, the robot achieved a 100\% success rate in more than 10 independent round-trip trials.
\end{abstract}


\section{Introduction}
\IEEEPARstart{I}{n}-pipe inspection is a representative constrained-environment robotics problem, where mobility, reliability, and compactness must be achieved under tight geometric and contact constraints \cite{choset2001coverage,nagatani2011redesign,song2015kinematic}. Practical pipelines contain elbows, diameter tolerances, and three-dimensional fittings that demand frequent posture correction; consequently, an in-pipe robot must concurrently deliver (i) robust axial propulsion and (ii) agile rolling reorientation for pose adaptation. In compact platforms, these two capabilities often interfere because they compete for limited contact resources inside the pipe, leading to slippage, jamming, or heavy feedback compensation. Beyond mobility, inspection operations often require systematic coverage and repeatable data acquisition for condition assessment and mapping; thus, locomotion repeatability and pose regulation directly affect the end-to-end inspection performance \cite{rayhana2020automated,xie2019automatic,li2019toward,macleod2016machining,kamezaki2025spira}.

A wide range of in-pipe robotic platforms have been explored \cite{verma2022review}, including tracked robots \cite{jang2022autonomous}, inchworm-like mechanisms \cite{liu2022tensegrity,di2025inchworm}, snake-like robots \cite{transeth2008snake,bando2016sound,virgala2021snake,ma2024locomotion}, and wheeled articulated designs such as V-shaped robots \cite{kakogawa2016design,kakogawa2024av}. Among them, V-shaped robots are attractive due to wheeled efficiency and multi-point contact. However, many V-shaped architectures remain kinematically coupled: rolling reorientation is difficult to generate as a pure, independently applied rolling torque because the central region is occupied by a co-located pivot/wheel shaft. As a result, rolling is often realized indirectly---for example, via differential wheel speeds \cite{kakogawa2018multi} or joint-pressure modulation \cite{kakogawa2024av}---and additional mechanisms (e.g., clutches) may be introduced to manage conflicts between traction and posture adjustment \cite{oka2021wheeled,dertien2014design}. These solutions increase mechatronic/control complexity and make performance sensitive to contact uncertainties.

\begin{figure}[t]
  \centering
  \includegraphics[width=0.35\textwidth]{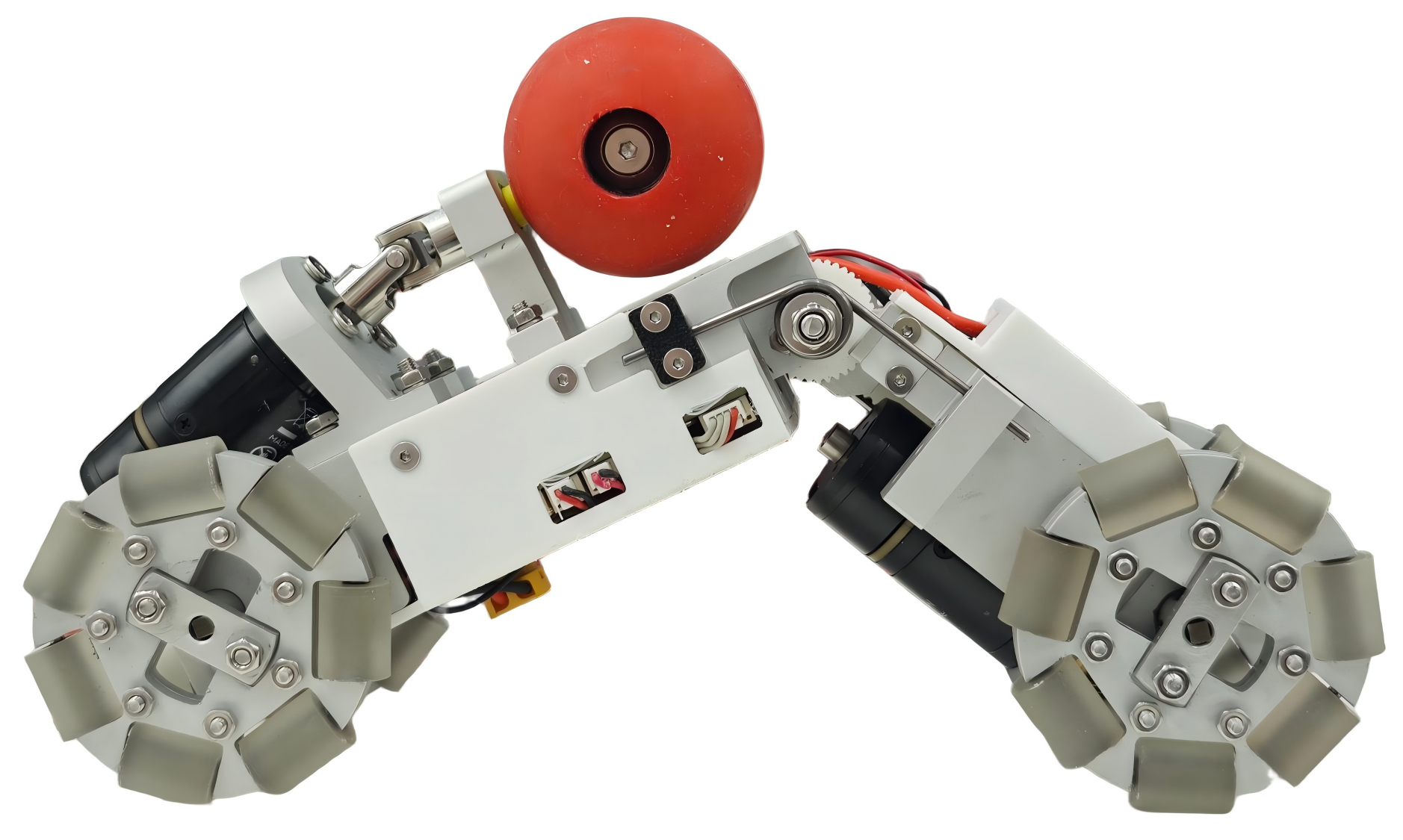}
  \caption{Prototype of the proposed V-shaped in-pipe robot enabled by a joint-axis-and-wheel-separation architecture.}
  \label{fig:Prototype}
\end{figure}

This paper adopts a structure-first perspective: we treat propulsion and rolling as a two-channel actuation problem and ask how the mechanical layout shapes the cross-channel transmission that a controller would otherwise need to compensate. Specifically, we describe the actuation process with a transmission matrix, where ideal decoupling corresponds to a near-diagonal mapping between commanded inputs (propulsion and rolling) and their intended outputs (axial traction and rolling reorientation). We show that, even when the propulsion and rolling actuators are spatially separated, the dominant residual off-diagonal term (roll-to-propulsion leakage) is set by the spherical-wheel contact geometry and can be parameterized by a single contact angle at the wheel--pipe interface. The contact angle determines both how much rolling torque is projected into undesired axial leakage and how efficiently the roll input is converted into useful rolling motion. This contact-angle-guided kinematic decoupling turns crosstalk suppression into a directly designable geometry problem: choose mounting parameters that drive the contact angle toward zero in the expected operating range. 
In this paper, kinematic decoupling refers to the near-diagonal actuation transmission from motor torques to body-level generalized torques (i.e., suppression of off-diagonal terms in the actuation transmission matrix).

Motivated by this insight, we propose the V-shaped in-pipe robot shown in Fig.~\ref{fig:Prototype}. The key architectural choice is a joint-axis-and-wheel-separation layout: the spherical-wheel shaft is spatially separated from the V-arm joint shaft, enabling a dedicated roll motor to apply an independent rolling torque while a dedicated propulsion motor drives all wheels for forward motion via an all-wheel-drive configuration. This separation converts the original coupled actuation problem into two largely independent channels, which is desirable for automation because it reduces the burden of cross-channel compensation and improves repeatability under varying contact conditions.

The contributions of this paper are threefold:
\begin{itemize}
  \item \textbf{A joint-axis-and-wheel-separation V-shaped in-pipe robot architecture} that achieves physically independent propulsion and rolling reorientation with only two motors, targeting agile posture correction in complex fittings.
  \item \textbf{A contact-angle-centric actuation transmission formulation} that makes the structure--performance link explicit: $\alpha$ serves as the geometric variable governing dominant roll-to-propulsion leakage and roll energy efficiency, providing a direct guideline for crosstalk suppression via mounting-parameter optimization.
  \item \textbf{Static-domain and experimental validation} showing (i) a stability-domain map for parameter selection under frictional uncertainty and (ii) hardware evidence of suppressed crosstalk via stable propulsion torque under high-dynamic rolling disturbances, together with repeatable traversal on a multi-material testbed including out-of-plane double elbows.
\end{itemize}

The remainder of this paper is organized as follows: Section~II presents the mechanical design and geometric constraints; Section~III establishes the static stability model and parameter selection guideline; Section~IV develops the actuation transmission model centered on the contact angle $\alpha$ and analyzes coupling suppression; Section~V reports experimental validation; and Section~VI concludes the paper.

\section{Mechanical Design and Geometric Constraints}
\label{sec:design}

\begin{figure}[t]
  \centering
  \includegraphics[width=0.32\textwidth]{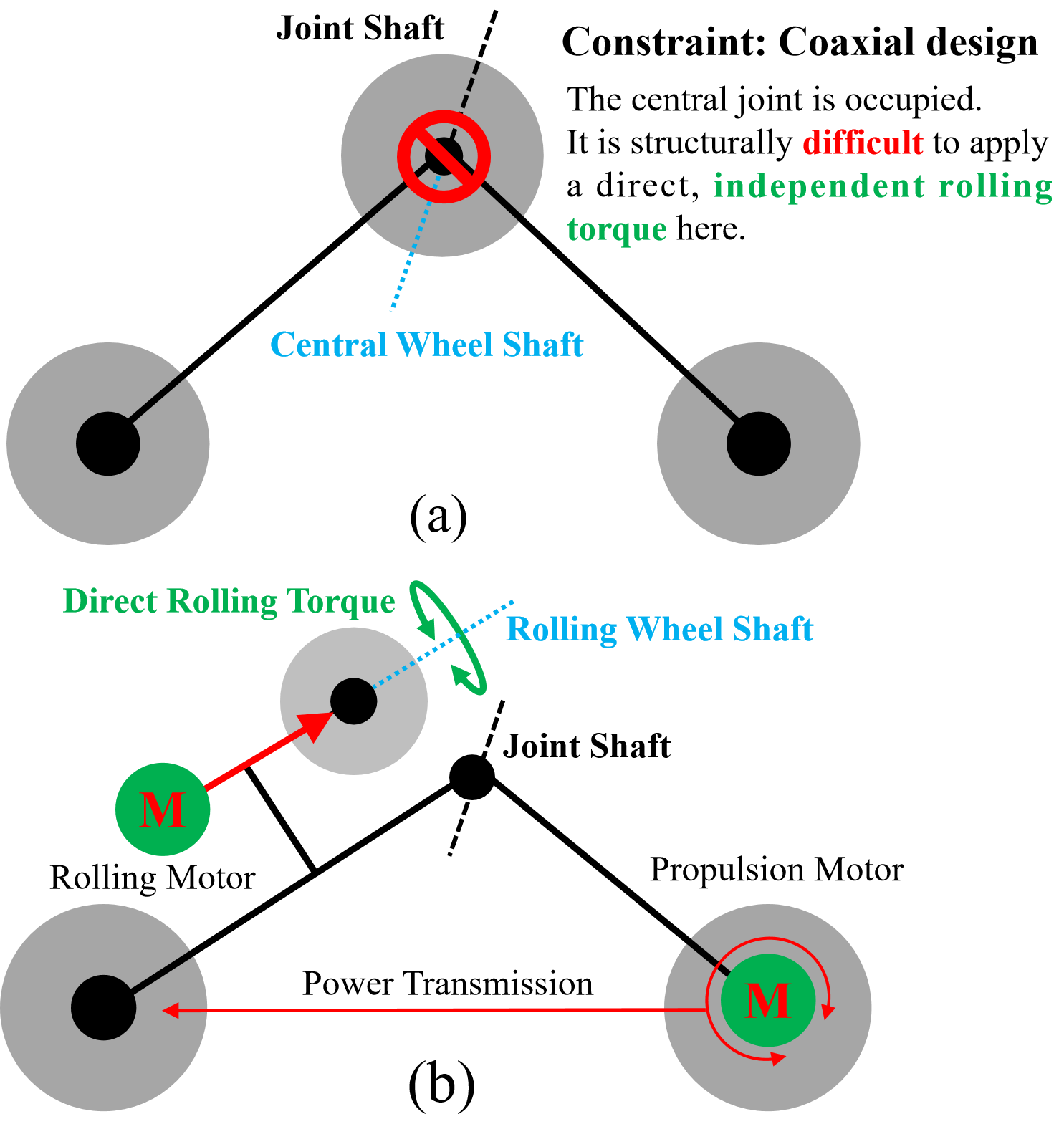}
  \caption{Conceptual comparison of architectures: (a) Coupled motion vs. (b) Our decoupled motion.}
  \label{fig:Decoupled}
\end{figure}

This section presents the mechanical architecture and the geometric configuration of the proposed V-shaped in-pipe robot.
The design targets two primary motions inside pipes: (i) axial propulsion and (ii) rolling reorientation about the pipe axis.
To reduce the intrinsic coupling between these motions that exists in conventional V-shaped robots \cite{kakogawa2024av,kakogawa2018multi,oka2021wheeled,dertien2014design},
we adopt a joint-axis-and-wheel-separation architecture (Fig.~\ref{fig:Decoupled}) and implement an all-wheel-drive (AWD) propulsion train combined with a centrally actuated spherical wheel for rolling control.

\begin{figure}[t]
  \centering
  \includegraphics[width=0.45\textwidth]{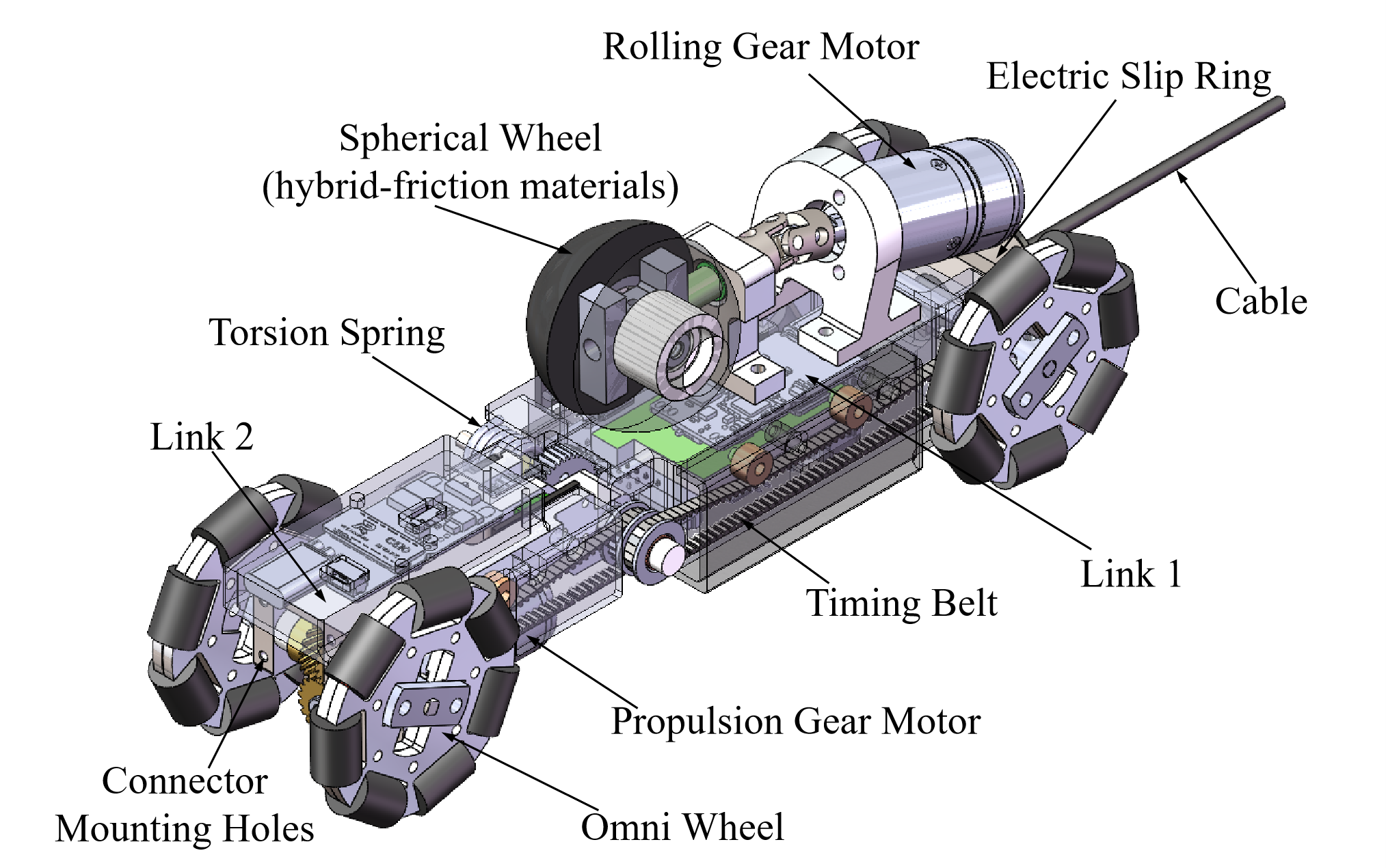}
  \caption{3D CAD model of the proposed in-pipe robot. The system integrates an AWD propulsion train and a centrally actuated spherical wheel for rolling reorientation.}
  \label{fig:CAD}
\end{figure}

\subsection{Decoupled Architecture and Actuation Allocation}
\label{subsec:arch}

Conventional V-shaped robots often co-locate the V-arm pivot axis and the central wheel shaft, which makes it difficult to apply a direct independent roll torque at the kinematic center.
Rolling reorientation is therefore commonly realized indirectly (e.g., differential wheel speeds or contact-pressure modulation) \cite{kakogawa2018multi,kakogawa2024av}, which increases control complexity and sensitivity to friction and geometry variations. To make this distinction explicit at the actuation-allocation level, representative Kakogawa-group designs are summarized in Table~\ref{tab:arch_comp}.

In contrast, the proposed architecture spatially separates the spherical-wheel shaft from the V-arm joint axis (Fig.~\ref{fig:Decoupled}).
This separation liberates the central region and enables a dedicated roll motor to apply a pure rolling torque through the spherical wheel,
while a separate propulsion motor drives all wheels for axial motion via an AWD transmission.
With only two motors, the system provides independent authority over propulsion and rolling reorientation, establishing the physical basis for the static and dynamic analyses in Sections~III and IV.


\begin{table*}[t!]
\centering
\begin{threeparttable}
\caption{Key architectural differences to representative close prior art V-shaped/articulated in-pipe robots at the actuation-allocation level.}
\label{tab:arch_comp}
\begin{tabularx}{\textwidth}{c c c c c >{\centering\arraybackslash}X}
\toprule
\textbf{Method} &
\textbf{P channel} &
\textbf{R channel} &
\textbf{P \& R co-available?} &
\textbf{\makecell{How the P--R conflict is resolved \\ (core mechanism)}} \\
\midrule
\textbf{Ours} &
\makecell{Dedicated AWD \\ propulsion motor} &
\makecell{Dedicated central\\ roll motor} &
\textbf{\makecell{Yes \\ (with two actuators)}} &
\makecell{\textbf{Physical separation} (joint-axis-and-wheel-separation) \\ enables direct center roll torque; \\ residual roll$\rightarrow$propulsion leakage is made \\ explicit and suppressible via geometry (contact angle)} \\
\midrule
\cite{oka2018stopper} (2018) &
\makecell{Shared input drives wheel  \\ \&  roll-joint \\ (underactuated differential)} &
\makecell{Emergent twisting/rolling \\ via load redistribution} &
\makecell{No \\ (alternated)} &
\makecell{\textbf{Time-multiplexing} by alternating front/rear wheel \\ driving + mechanical stopper limiting roll-joint motion} \\
\midrule
\cite{kakogawa2019pipe} (2019) &
High-traction mode &
\makecell{Rolling mode \\ (direction-dependent)} &
\makecell{No \\ (mode-dependent)} &
\makecell{\textbf{Mode switching} by middle-joint rotation direction; \\ rolling is largely \textbf{contact-mediated} (environment reaction) \\ rather than an independent commanded roll-torque channel } \\
\midrule
\cite{oka2022control} (2021) &
\makecell{Underactuated + one-way \\ clutch constraint} &
\makecell{Mode-select \\ via clutch/direction} &
\makecell{No \\ (mode-switch)} &
\makecell{\textbf{Clutch-constrained separation}: switching between \\ motion modes by selecting motor drive direction \\ (not simultaneous independent P/R)} \\
\midrule
\cite{kakogawa2024av} (2024) &
Multi-DOF actuation &
Multi-DOF actuation &
\textbf{\makecell{Yes \\ (with six actuators)}} &
\makecell{\textbf{Actuation redundancy}: reduces conflict mainly \\ by adding DOFs/actuators \\ (higher mechatronic/control complexity)} \\
\bottomrule
\end{tabularx}
\begin{tablenotes}[flushleft, online] 
\footnotesize
\item[\textbf{Note:}] P: axial propulsion (traction); R: rolling reorientation about the pipe axis.
\end{tablenotes}
\end{threeparttable}
\end{table*}

\subsection{Prototype Implementation and Robustness Mechanisms}
\label{subsec:impl}

Fig.~\ref{fig:CAD} shows the 3D CAD model of the realized prototype.
The propulsion subsystem uses four side omni-wheels driven in an AWD manner through a synchronous belt.
The passive rollers on the omni-wheels allow low-resistance coordination while maintaining robust traction redundancy.
The rolling subsystem uses a centrally actuated spherical wheel at a persistent contact location, which provides fast posture adjustment even when some propulsion contacts are temporarily degraded.

A preloaded torsion spring at the V-arm joint provides passive compliance, enabling adaptation to pipe diameter variations without active force regulation.
In addition, the spherical wheel is engineered with a hybrid-friction interface to realize a fault-tolerant functional slip behavior, which converts a potential high-resistance lock-up near unfavorable contact states into a predictable axial slip that can be compensated by the AWD propulsion.

\subsection{Geometric Constraints, Configuration Analysis, and Contact Angle Definition}
\label{subsec:geom_alpha}

\begin{figure}[t]
  \centering
  \includegraphics[width=0.48\textwidth]{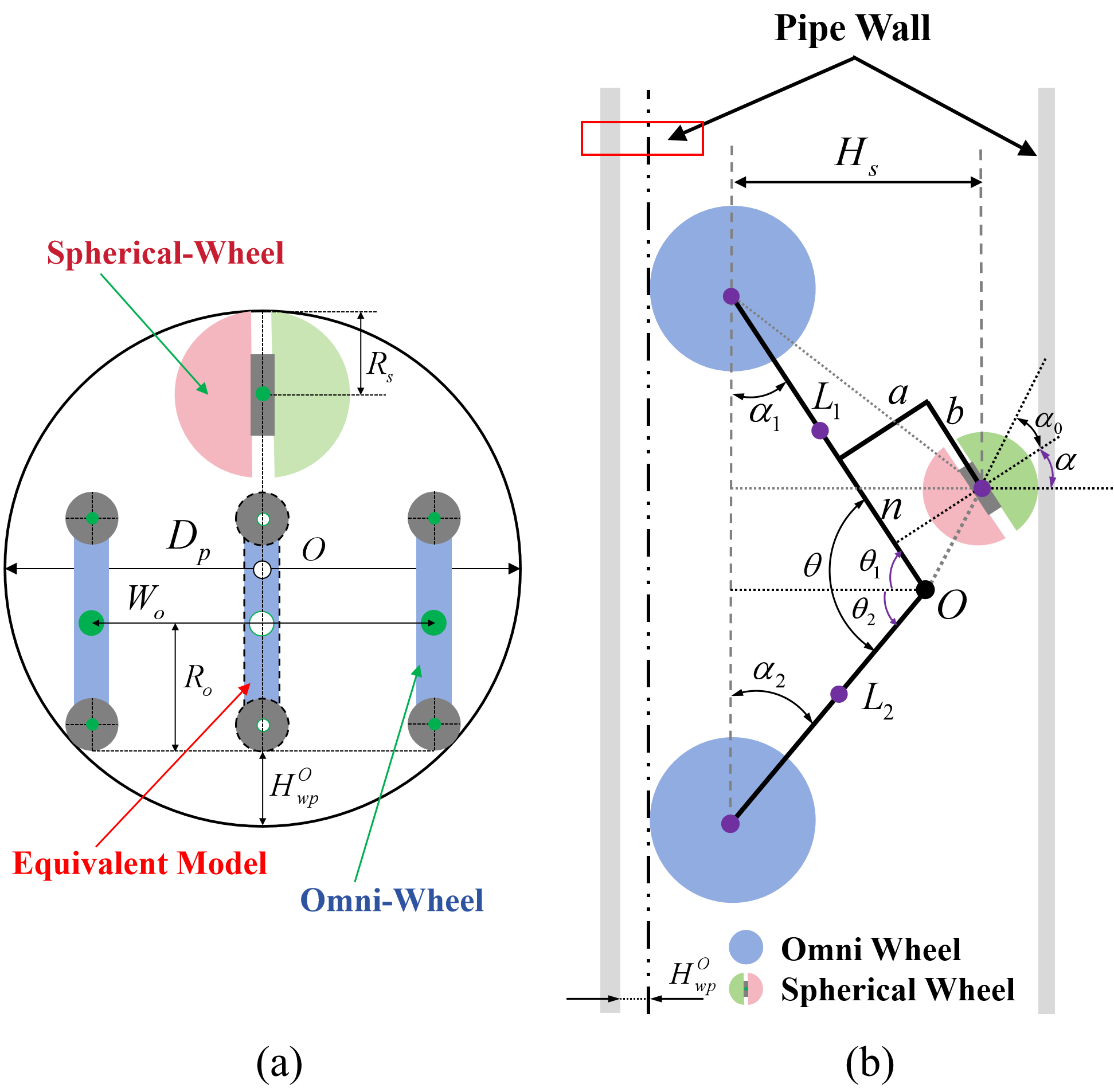}
  \caption{
  Geometric constraints and end-view configuration of the V-shaped robot inside a circular pipe. 
  (a) Pipe cross-section schematic defining the pipe diameter $D_p$, the omni-wheel pair spacing $W_o$, the wheel radii $R_o$ (omni wheel) and $R_s$ (spherical wheel), the pipe center $O$, and the nominal omni-wheel-to-wall geometric clearance $H_{wp}^{0}$ shown in the equivalent 2-D model. 
  (b) End-view configuration used for the geometric constraint and contact-angle analysis: the core structural design parameters $(L_1, L_2, a, b, n)$ determine the derived configuration angles $(\alpha_1,\alpha_2,\theta_1,\theta_2)$ and the total V-opening angle $\theta=\theta_1+\theta_2$. 
  The distance $H_s$ denotes the offset between the omni-wheel centerline and the spherical-wheel center along the pipe diameter direction (as indicated). 
  At the spherical-wheel--pipe interface, $\alpha_0$ marks the angular location of the nominal (ideal) contact point on the pipe wall, while $\alpha$ denotes the actual contact angle used later to characterize transmission crosstalk.}
  \label{fig:GeoConstraint}
\end{figure}

\begin{figure}[t]
  \centering
  \includegraphics[width=0.44\textwidth]{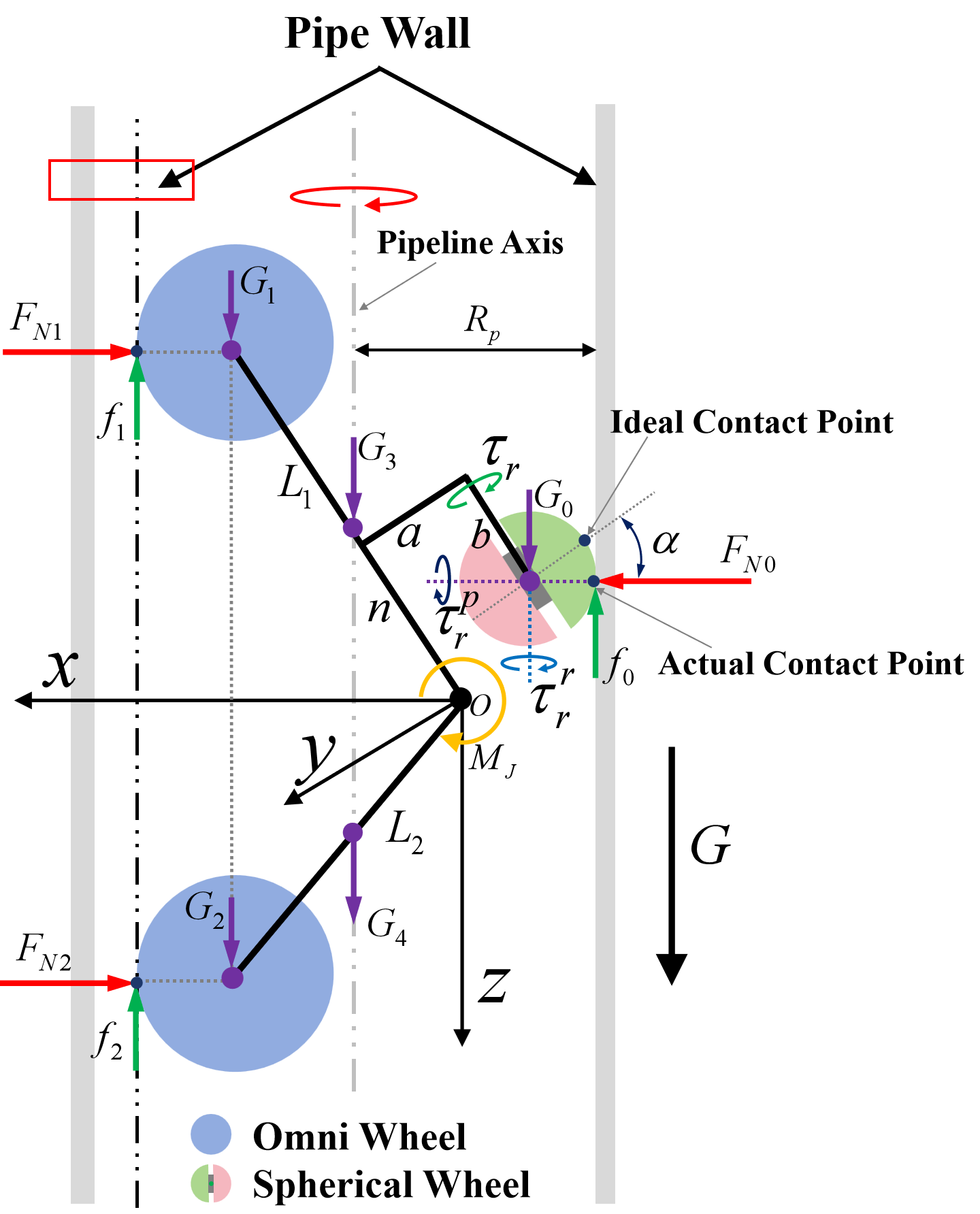}
  \caption{
  Unified end-view model and notation used in both the static stability analysis (Section~III) and the actuation transmission analysis (Section~IV). 
  The model defines the wheel--pipe normal forces $F_{N0},F_{N1},F_{N2}$ at the spherical wheel (index 0) and the two omni-wheel pairs (indices 1 and 2), the corresponding friction forces $f_0,f_1,f_2$, the gravitational loads $G_k$ of the main modules (indexed in the figure), and the joint torque $M_J$ at the V-arm pivot. 
  The inset illustrates the spherical-wheel contact and defines the contact angle $\alpha$ as the deviation between the direction of the actual contact force/velocity transmission and the ideal circumferential direction for pure rolling about the pipe axis; the rolling torque $\tau_r$ and the induced axial-leakage component $\tau_p$ are also illustrated.}
  \label{fig:AlphaDef}
\end{figure}

To make the structure--performance relationship explicit, we analyze the robot end-view geometry constrained inside a circular pipe (Fig.~\ref{fig:GeoConstraint}) and parameterize the configuration using a minimal set of structural design parameters $(L_1, L_2, a, b, n, W_o)$ together with pipe constraints $(D_p)$. Based on this configuration, we define the spherical-wheel contact angle $\alpha$ at the wheel--pipe interface, which will be used throughout the paper as the key geometric variable linking the mechanical layout to (i) static force allocation and stability margin (Section~III) and (ii) roll--propulsion transmission crosstalk (Section~IV). All symbols follow the definitions in Fig.~\ref{fig:GeoConstraint}(b) and Fig.~\ref{fig:AlphaDef}.

As illustrated in Fig.~\ref{fig:GeoConstraint}(a), the omni-wheel pair has a spacing $W_o$ inside a pipe of diameter $D_p$.
The geometric clearance between the omni-wheel and the pipe wall is
\begin{equation}
H_{wp}^{o}=\frac{1}{2}\left(D_p-\sqrt{D_p^{2}-W_o^{2}}\right).
\label{eq:Howp}
\end{equation}
Let $R_o$ and $R_s$ denote the radii of the omni-wheel and the spherical wheel, respectively.
Then the distance between the centerline of the omni-wheels and the center of the spherical wheel (Fig.~\ref{fig:GeoConstraint}) is
\begin{equation}
H_s = D_p - R_s - H_{wp}^{o} - R_o .
\label{eq:Hs}
\end{equation}

As shown in Fig.~\ref{fig:GeoConstraint}(b), the configuration of the V-shaped arms is determined by the core structural parameters $L_1, L_2, a, b, n$.
From the geometric relationships within Link~1 ($L_1$), the angle $\alpha_1$ between Link~1 and the horizontal line is derived as
\begin{equation}
\alpha_1
=
\sin^{-1}\!\left(\frac{H_s}{\sqrt{(L_1-n+b)^2+a^2}}\right)
-
\tan^{-1}\!\left(\frac{a}{L_1-n+b}\right).
\label{eq:alpha1}
\end{equation}
From the triangle formed by the centers of the two wheels, the corresponding angle for Link~2 ($L_2$) is obtained as
\begin{equation}
\alpha_2 = \sin^{-1}\!\left(\frac{L_1\sin\alpha_1}{L_2}\right).
\label{eq:alpha2}
\end{equation}

These angles determine the arm orientations and thus the V-arm configuration. We define the configuration angles as
$\theta_1 = \frac{\pi}{2}-\alpha_1$ and $\theta_2 = \frac{\pi}{2}-\alpha_2$, and the total V-opening angle is $\theta=\theta_1+\theta_2$ (Fig.~\ref{fig:GeoConstraint}(b)). 

In later sections, we characterize the non-ideal spherical-wheel contact by the deviation between the resultant contact force/velocity transmission direction and the ideal circumferential direction required for pure rolling about the pipe axis (Fig.~\ref{fig:AlphaDef}, inset). We denote this deviation as the contact angle $\alpha$. Under the end-view kinematic constraint, the ideal circumferential direction is orthogonal to the omni-wheel centerline, which leads to a purely geometric expression
\begin{equation}
\alpha = \frac{\pi}{2}-\theta_1 = \alpha_1,
\label{eq:alpha_def}
\end{equation}
where principal values are taken such that $\alpha,\alpha_1\in[0,\pi/2]$ for the physical configuration considered.

\subsection{Prototype Specifications}
\label{subsec:spec}

Table~\ref{tab:spec} summarizes the key specifications of the developed prototype.

\begin{table}[t]
  \centering
  \caption{Key specifications of the robot module}
  \label{tab:spec}
  \begin{tabular}{l c}
    \hline
    \textbf{Specification} & \textbf{Value} \\
    \hline
    Length of Link 1 ($L_1$) & 105 mm \\
    Length of Link 2 ($L_2$) & 75 mm \\
    Total Length & 240 mm \\
    Total Mass & 750 g \\
    Omni-wheel Radius ($R_o$) & 30 mm \\
    Spherical-wheel Radius ($R_s$) & 22.5 mm \\
    Applicable Pipe Diameter & 100 mm \\
    Torsion Spring Stiffness & 10.06 $\textup{N}\cdot\textup{mm}$/deg\\
    \hline
  \end{tabular}
\end{table}

\section{ STATIC MODELING AND STABILITY ANALYSIS}
This section establishes a static model of the robot to address two objectives that are prerequisite to the subsequent transmission-crosstalk analysis. First, it evaluates the robot’s anti-gravity feasibility in a vertical pipe by deriving a stability condition under friction uncertainty. Second, it computes the distribution of wheel--pipe normal forces at all contact points. These normal forces directly determine the available friction envelopes for both axial propulsion and rolling reorientation, and will be used as the physical basis for the dynamic actuation transmission analysis in Section~IV.

\subsection{Static Equilibrium Model} 
This subsection establishes a static equilibrium model of the robot in a vertical pipe and derives the wheel--pipe normal forces under gravity loading. The geometric constraints and configuration relationships (including the definition of the contact angle and configuration angles) have been given in Section~II-C. Based on that configuration, we formulate force/moment equilibrium to solve the normal forces and then evaluate the anti-gravity condition using friction limits. The explicit closed-form solution and the moment-arm definitions used in the equilibrium equations are provided in Appendix~A for completeness.

As shown in Fig.~\ref{fig:AlphaDef}, we use an equivalent 2-D planar static model to capture the contact geometry and load transfer for a vertical-pipe scenario. The structural parameters $(L_1,L_2,a,b,n)$ determine the configuration (Section~II-C), and the static model solves three unknown normal forces at the wheel--pipe contacts, $F_{N0}$ (spherical wheel) and $F_{N1},F_{N2}$ (two omni-wheel pairs). The external loads are the component weights $G_k$ (indexed in Fig.~5), and the joint provides a torsion-spring torque $M_J$ that sets the clamping level. In this subsection we focus on obtaining $F_{Ni}$; friction forces are subsequently accounted for through their upper bounds $\mu_i F_{Ni}$ when evaluating the anti-gravity condition in Section~III-B2.

We employ a subsystem moment-balance method by virtually ``cutting'' the V-shaped structure at the joint $O$ (Fig.~5). Force equilibrium in the lateral direction provides one scalar equation, and two additional scalar equations are obtained by taking moments of the two subsystems (Link~1 side and Link~2 side) about the joint $O$. All moments are computed using vector moment arms referenced to $O$, and the counter-clockwise direction is defined as positive. Because the torsion spring applies equal and opposite torques to the two arms, the joint torque $M_J$ enters the two subsystem moment equations with opposite signs.

Overall lateral force balance ($\Sigma F_x=0$):
\begin{equation}
F_{N1}+ F_{N2}- F_{N0}=0
\label{eq: eq1}
\end{equation}
Moment balance for subsystem 1 (Link 1 side) about point O ($\Sigma M^{L_1}_{o}=0$):
\begin{equation}
M_{N0} + M_{N1} + M_{G0}+ M_{G1}+ M_{G3}- M_J = 0
\label{eq: eq2}
\end{equation}
Moment balance for subsystem 2 (Link 2 side) about point O ($\Sigma M^{L_2}_{o}=0$):
\begin{equation}
M_{N2}  + M_{G2}+ M_{G4}+ M_J = 0
\label{eq: eq3}
\end{equation}
where $M_{Ni}$ denotes the moment about $O$ generated by the normal force $F_{Ni}$ at contact $i$ ($i=0,1,2$), and $M_{Gk}$ denotes the moment about $O$ generated by the weight $G_k$ of component $k$ (indexed in Fig.~\ref{fig:AlphaDef}). By solving \eqref{eq: eq1} - \eqref{eq: eq3} simultaneously, we obtain analytical expressions of $F_{Ni}$ in terms of the design parameters and the pipe constraints. The explicit closed-form expressions and the corresponding moment-arm definitions are provided in Appendix~A.

\subsection{Results and Discussion}
\subsubsection{\textbf{Effect of Structural Parameters on Normal Force Distribution}}

\begin{figure}[t]
  \centering
  \includegraphics[width=0.42\textwidth]{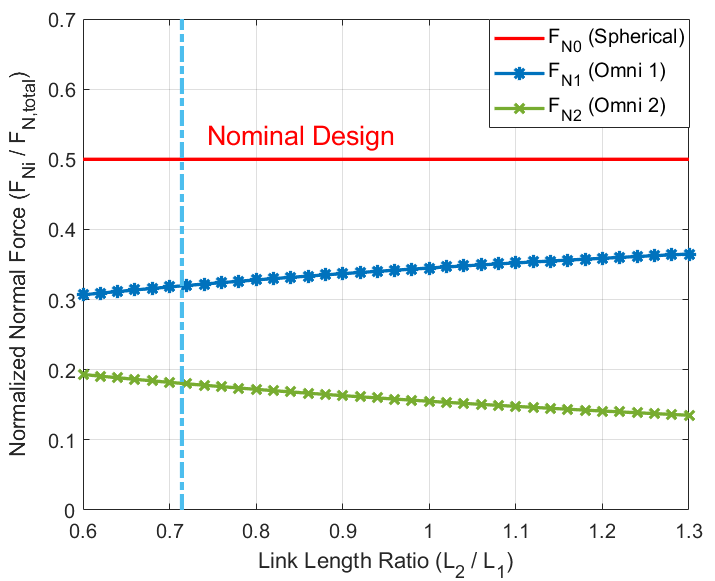}
  \caption{  Effect of link length ratio on normal force distribution. In this analysis, the length of link $L_1$ is fixed at 105 mm, and different $L_2/L_1$ ratios are obtained by varying the length of link $L_2$. All other design parameters are taken from the nominal values in Table~\ref{tab:spec}. The gray dashed lines represent the length of link actual values in Table~\ref{tab:spec}.}
  \label{fig: Force-VS-K10}
\end{figure}

To reveal how structural asymmetry affects the internal load sharing, we use the static model to analyze the influence of a key dimensionless parameter---the link length ratio $L_2/L_1$---on the normal-force distribution, as shown in Fig.~\ref{fig: Force-VS-K10}. In this analysis, $L_1$ is fixed at $105~\mathrm{mm}$ and $L_2$ is varied to obtain different $L_2/L_1$ ratios, while all other design parameters are set to their nominal values in Table~\ref{tab:spec}. The resulting $F_{Ni}$ distribution provides direct guidance on how much contact pressure (and thus friction capacity) can be allocated to each wheel group.

A notable characteristic in Fig.~\ref{fig: Force-VS-K10} is that the spherical-wheel normal force $F_{N0}$ remains close to $50\%$ of the total normal force $F_{N,\mathrm{total}}=F_{N0}+F_{N1}+F_{N2}$ over the entire range of $L_2/L_1$. This indicates that the V-shaped contact topology intrinsically assigns a dominant clamping reaction to the spherical wheel, while the arm-length asymmetry primarily redistributes the remaining reaction between $F_{N1}$ and $F_{N2}$. As $L_2/L_1$ increases, the share of $F_{N1}$ increases and that of $F_{N2}$ decreases, which directly impacts the available friction capacity on Link~2.

This analysis provides quantitative guidance for structural design. 
To prevent the omni-wheel on Link 2 ($F_{N2}$) from slipping due to insufficient contact pressure, the $L_2/L_1$ ratio should not be excessively large. Our current design ($L_2/L_1 \approx  0.71$) is a choice made based on practical engineering trade-offs, ensuring that $F_{N2}$ bears nearly 20\% of the total normal force, providing effective multi-point support for the system. Furthermore, the AWD configuration ensures that an effective driving force can still be allocated to wheel 1 ($F_{N1}$), even when its pressure is not the maximum. This improves the robot's propulsive performance and reduces the risk of failure caused by wheel slip due to insufficient pressure.

\subsubsection{\textbf{Stability Domain Analysis under Frictional Uncertainty}}
To evaluate robustness under frictional uncertainty in real pipelines, we conduct a stability-domain analysis for a vertical pipe. The anti-gravity condition is that the maximum total static friction that the robot can generate along the pipe axis is no less than its total weight. We define the stability safety margin as
$S = f_{\max,\mathrm{total}}/G_{\mathrm{total}}$, where
$f_{\max,\mathrm{total}}$ is the total available static-friction upper bound from all contacts. The system is stable against gravity if $S\ge 1$, i.e.,
\begin{equation}
\mu_s F_{N0}+\mu_o(F_{N1}+F_{N2}) \ge G_{\mathrm{total}} .
\label{eq: sliding}
\end{equation}

Here, $F_{Ni}$ are the normal forces solved from \eqref{eq: eq1} - \eqref{eq: eq3}, which already include the effect of gravity through the weight moments. $\mu_o=0.3$~\cite{kakogawa2018design} is the effective axial friction coefficient for the omni-wheel contact (accounting for the roller-based contact characteristics in pipe environments), and $\mu_s$ is the static friction coefficient at the spherical-wheel--pipe interface, which may vary with hybrid-friction materials and surface conditions. Equation~\eqref{eq: sliding} therefore provides a conservative upper bound on the anti-gravity capability for a given normal-force distribution.

\begin{figure}[t]
  \centering
  \includegraphics[width=0.48\textwidth]{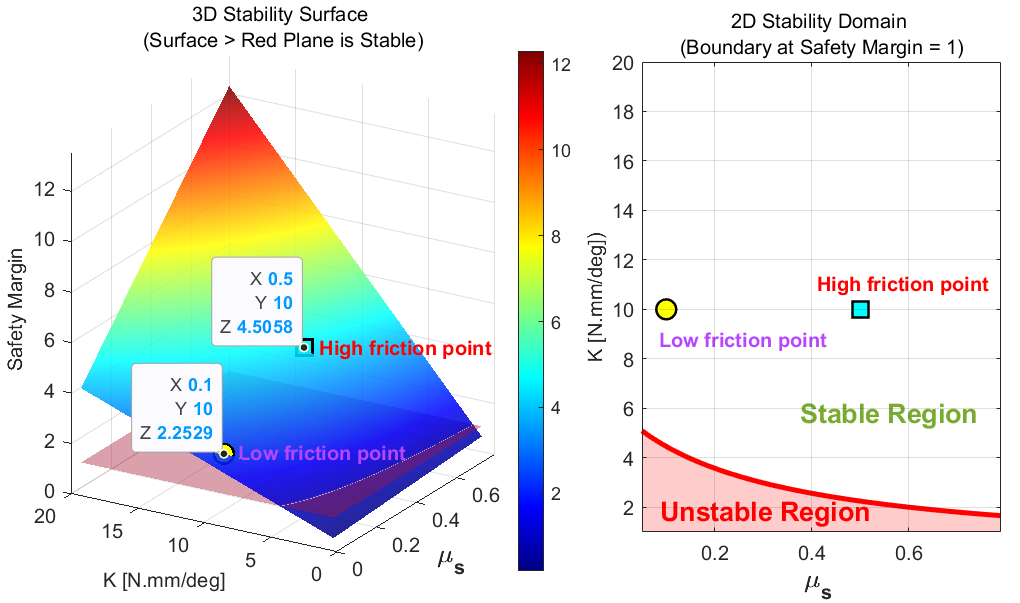}
  \caption{Stability-domain analysis showing the effect of torsion-spring stiffness $K$ and spherical-wheel friction coefficient $\mu_s$ on anti-gravity feasibility. (a) Three-dimensional stability surface, where the stability safety margin is defined as $S=f_{\max,\mathrm{total}}/G_{\mathrm{total}}$. When the surface is above the red reference plane $S=1$, the system is stable against gravity in a vertical pipe. (b) Two-dimensional stability-domain plot, obtained by projecting (a) onto the $(\mu_s, K)$ plane. The white region indicates stable operation ($S\ge 1$) and the red region indicates slipping ($S<1$). The thick red curve is the critical boundary at $S=1$, with $K$ reported in $\mathrm{N\cdot mm/deg}$.}
  \label{fig: 3D-SP-K20}
\end{figure}

We plot the critical stability boundary implied by \eqref{eq: sliding} in the design space spanned by the spherical-wheel friction $\mu_s$ and the torsion-spring stiffness $K$, resulting in the stability map shown in Fig.~\ref{fig: 3D-SP-K20}. The white region corresponds to $S\ge 1$ (stable against gravity), whereas the red region corresponds to $S<1$ (slipping). This map provides a direct guideline for selecting $K$: given an expected lower bound of $\mu_s$ in the target environment, choose the smallest $K$ such that the operating point lies inside the stable region with a desired margin (e.g., $S\ge 1.5$), which ensures anti-gravity feasibility while avoiding unnecessarily large clamping forces.

Fig.~\ref{fig: 3D-SP-K20} illustrates the stability domain and confirms the trade-off between torsion-spring stiffness $K$ and spherical-wheel friction $\mu_s$: a larger $K$ can compensate for a smaller $\mu_s$ by increasing the normal forces and thus the friction upper bound in \eqref{eq: sliding}. Importantly, the stiffness value used in our prototype is not a virtual parameter. The installed torsion spring is characterized by
$K = 1025.85~\mathrm{g\cdot mm/deg}$, which corresponds to
$K \approx 10.06~\mathrm{N\cdot mm/deg}$ using $1~\mathrm{g\cdot mm}=9.80665\times 10^{-3}~\mathrm{N\cdot mm}$.
For consistency with Fig.~7, we report $K$ in $\mathrm{N\cdot mm/deg}$ throughout this section. With $\mu_s=0.1$, the nominal operating point $(K\approx 10.06~\mathrm{N\cdot mm/deg},\,\mu_s=0.1)$ lies securely inside the stable region with a safety margin of $S=2.3$, leaving a $56.5\%$ traction reserve against gravity (see Appendix~B for the definition).

In conventional coupled designs, clamping force, propulsion, and rolling are often generated by the same actuation pathway, which makes the achievable contact forces and motion capabilities strongly coupled and sensitive to parameter variations. In contrast, our architecture allows the clamping level to be set primarily by the passive torsion spring (through $K$) while keeping propulsion and rolling as two independent active tasks. Selecting $K$ within the stable domain yields a predictable normal-force distribution and a guaranteed anti-gravity margin, which provides a well-defined physical operating condition for the actuation-transmission analysis and dynamic evaluation in Section~IV.

\section{DYNAMIC MODELING AND COUPLING ANALYSIS}
This Section aims to establish a comprehensive dynamic model to quantitatively reveal how the design parameters of our proposed ``wheel-joint axis separation" architecture influence its decoupling performance. We first develop a unified dynamic model that integrates the robot's rigid-body dynamics with the transmission characteristics of its actuation system. Subsequently, we conduct a physical-level modeling of the actuation system to analyze its coupling coefficients and energy efficiency. Finally, simulation-based validation demonstrates the decisive impact of structural optimization of the contact angle on the system's dynamic performance.

\subsection{Dynamic Model and Actuation Transmission Model}
The macroscopic dynamic behavior of the robot can be described by the general equation of motion derived using the Lagrangian method:
\begin{equation}
{\mathbf{M(q)\ddot{q}+C(q,\dot{q})\dot{q}+G(q) = Q_{drive}}}
\label{eq:D_Q}
\end{equation}
where $\mathbf{q}=[z(t),\phi(t)]^{T}$ is the generalized coordinate vector with $z$ the axial displacement and $\phi$ the roll angle, and $\mathbf{Q_{drive}}$ is the vector of generalized forces applied to the robot body by the actuation system. As shown in Appendix~\ref{app:inertial_crosstalk}, the ideal rigid-body inertia matrix of a V-shaped structure has zero off-diagonal terms, implying that the dominant coupling effects in practice originate from the transmission process of the generalized force vector rather than inertial translation--rotation coupling. Therefore, accurately modeling $\mathbf{Q_{drive}}$ is key to understanding the system's dynamic performance.

To characterize physical-layer crosstalk, we define two core vectors:
\begin{itemize}
    \item The commanded motor-shaft torque vector
    \begin{equation}
    \mathbf{\tau}_{\mathrm{cmd}} = [\tau_{p},\,\tau_{r}]^{T},
    \label{eq:tau_cmd}
    \end{equation}
    where $\tau_{p}$ and $\tau_{r}$ are the commanded torques of the propulsion motor and the roll motor, respectively.
    \item The body-level generalized output-torque vector applied by the actuation system
    \begin{equation}
    \mathbf{Q}_{\mathrm{drive}} = [M_{p},\,M_{r}]^{T},
    \label{eq:Q_drive}
    \end{equation}
    where $M_p$ is the equivalent propulsive torque defined as $M_p = F_p\,R_o$ ($F_p$: \textbf{total axial propulsive force}; $R_o$: omni-wheel radius), and $M_r$ \textbf{is the equivalent rolling torque} about the pipe axis applied to the robot body.
\end{itemize}

The mapping between motor commands and body-level generalized torques is described by a 2$\times$2 actuation transmission matrix:
\begin{equation}
\mathbf{Q}_{\mathrm{drive}} = \mathbf{T}_{A}\,\mathbf{\tau}_{\mathrm{cmd}},
\label{eq:T_A}
\end{equation}
where $\mathbf{T}_{A}$ is determined by the mechanical layout and transmission principles:
\begin{equation}
\mathbf{T}_{A}=
\left[
\begin{array}{cc}
k_{pp} & k_{pr}\\
k_{rp} & k_{rr}
\end{array}
\right].
\label{eq:T_de}
\end{equation}
All elements of $\mathbf{T}_{A}$ are dimensionless because both input and output channels are expressed in torque units. The diagonal terms ($k_{pp}$, $k_{rr}$) represent the primary transmission gains of propulsion and rolling channels, while the off-diagonal terms ($k_{pr}$, $k_{rp}$) quantify leakage between channels. In our architecture, the dominant, design-governed leakage is the roll-to-propulsion term $k_{pr}$, which is directly controlled by the spherical-wheel contact angle $\alpha$; $k_{rp}$ is secondary and mainly arises from asymmetry and friction non-idealities.

\subsection{Physical Modeling of the Actuation Transmission Matrix and Contact Angle Analysis}
The coefficients of $\mathbf{T}_{A}$ are determined by the physical transmission mechanisms of the two actuation channels. In particular, the rolling channel is governed by the contact interaction between the central spherical wheel and the pipe wall. As illustrated in Fig.~\ref{fig:AlphaDef}, when the roll motor drives the spherical wheel, the contact traction is generally not aligned with the ideal circumferential direction required for pure rolling about the pipe axis. We characterize this directional deviation by the contact angle $\alpha$. The available traction magnitude is proportional to the normal force $F_{N0}$, whose static distribution and dependence on structural parameters were analyzed in Section~III. Under the nominal straight-pipe/full-contact configuration, the geometric constraints in Section~II (Eqs.~\eqref{eq:Howp}--\eqref{eq:alpha_def}) yield $\alpha$ (Eq.~\eqref{eq:alpha_def}) as a function of the mounting parameters. In pipe bends the local contact geometry can vary; nevertheless, minimizing $|\alpha|$ in the nominal configuration provides a physically grounded design handle that robustly suppresses the dominant leakage term $|k_{pr}|\propto|\sin\alpha|$, as further supported by the experimental results.

Because the contact between the spherical wheel and the pipe wall is non-ideal, a deviation exists between the direction of the actual contact force and the ideal circumferential direction. We define this deviation as the contact angle $\alpha$. Under the assumption of full contact in a straight pipe, the geometric constraints and configuration analysis in section II (Eqs.~\eqref{eq:Howp}--\eqref{eq:alpha_def}) yield $\alpha$ as a function of the robot's structural parameters:

\subsubsection{Roll Channel}
The contact angle $\alpha$ projects the roll-motor input torque $\tau_{r}$ into two components: an effective rolling component $\tau_{r}^{r}$ and a leakage-induced propulsive component $\tau_{r}^{p}$. This projection can be expressed as
\begin{equation}
\tau_{r}^{r}=\tau_{r}\cos\alpha,\qquad
\tau_{r}^{p}=\tau_{r}\sin\alpha.
\label{eq:tau_r}
\end{equation}
Assuming the robot’s center of rotation during a roll maneuver approximates the pipe central axis, the lever arm for rolling is the pipe radius $R_p=D_p/2$. Consequently, the rolling-channel gain $k_{rr}$ and the roll-to-propulsion leakage coefficient $k_{pr}$ are
\begin{equation}
k_{rr}=(R_p/R_s)\cos\alpha,\qquad
k_{pr}=(R_o/R_s)\sin\alpha,
\label{eq:k_r}
\end{equation}
where $R_s$ is the spherical-wheel radius and $R_o$ is the omni-wheel radius.

\subsubsection{Propulsion Channel}
The primary transmission gain of the propulsion channel, $k_{pp}$, is determined by the AWD drivetrain torque amplification and the number of driven wheels:
\begin{equation}
k_{pp}=G_p\,N,
\label{eq:k_p}
\end{equation}
where $G_p$ is the effective torque gain from the propulsion motor shaft to each driving omni wheel (including gearing), and $N$ is the number of driven omni wheels. The propulsion-to-roll leakage term $k_{rp}$ mainly arises from non-idealities such as structural asymmetry and friction anisotropy that can introduce a small net circumferential traction imbalance during axial driving. In our design, this effect is secondary because the omni-wheel structure suppresses unintended circumferential traction and the normal-force distribution remains bounded and predictable (Section~III). Therefore, the dominant, geometry-governed leakage term is $k_{pr}$, and the decoupling optimization focuses on minimizing $|k_{pr}|$ through contact-angle reduction.

\subsubsection{Energy Efficiency}
A physically consistent proxy for posture-adjustment efficiency is the fraction of the roll motor torque that contributes to useful rolling about the pipe axis. For a fixed rolling displacement (i.e., a fixed angular stroke), the mechanical work scales linearly with the effective torque component. Using the projection in \eqref{eq:tau_r}, we define the roll actuation efficiency as
\begin{equation}
\eta_{\mathrm{roll}}=\tau_{r}^{r}/\tau_{r}=\cos\alpha.
\label{eq:eta_roll}
\end{equation}
This expression shows that increasing $|\alpha|$ (e.g., due to unfavorable mounting or local contact changes in fittings) reduces rolling authority and increases the parasitic component that must be counteracted by the propulsion channel to maintain axial position. Therefore, minimizing $|\alpha|$ is beneficial for both suppressing leakage $|k_{pr}|\propto|\sin\alpha|$ and preserving rolling effectiveness.

\subsection{Analysis of Structural Parameter Effects on Coupling Characteristics}
\begin{figure}[thpb]
  \centering
  \includegraphics[width=0.4\textwidth]{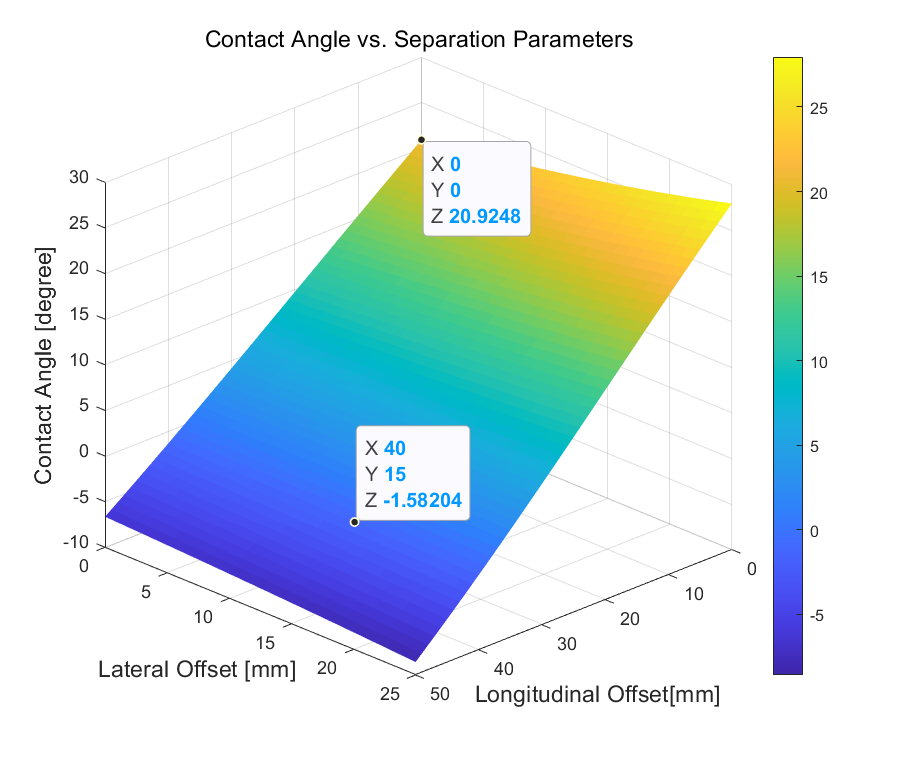}
  \caption{Contact-angle surface $\alpha(a,\,n-b)$ computed from the geometric constraints (Eq.~\eqref{eq:alpha_def}). The marked point corresponds to the prototype mounting parameters.}
  \label{fig:alpha_3d_png}
\end{figure}

This section quantitatively analyzes how the spatial separation of the central spherical wheel affects coupling characteristics. The separation is parameterized by the mounting variables $a$, $b$, and $n$. We define a configuration with $a=b=n=0$ as a non-optimized baseline within the same architecture (“zero separation”), in which the actuation unit is placed directly at the pivot axis. Using Eq.~\eqref{eq:alpha_def}, we compute the contact angle $\alpha$ as a function of the mounting parameters $a$ (longitudinal offset) and $n-b$ (lateral offset), and visualize it in Fig.~\ref{fig:alpha_3d_png}.

To visually demonstrate the decisive impact of spatial separation on system performance, according to \eqref{eq:alpha_def}, we can plot the contact angle $\alpha$ as a function of the mounting parameters $a$ (longitudinal offset) and $n-b$ (lateral offset), as shown in the performance surface plot in Fig. \ref{fig:alpha_3d_png}. 

This plot reveals several key design principles:
\begin{itemize}
    \item \textbf{Zero separation yields non-negligible crosstalk potential:}
    At the origin ($a=0,\, n-b=0$), the baseline contact angle magnitude is relatively large (e.g., $|\alpha|\approx 21^\circ$). According to \eqref{eq:k_r}, this increases the leakage magnitude $|k_{pr}|\propto|\sin\alpha|$ and reduces the rolling effectiveness $\eta_{\mathrm{roll}}=\cos\alpha$ (e.g., $\eta_{\mathrm{roll}}\approx 0.93$).
    \item \textbf{Spatial separation reduces $|\alpha|$ and thus suppresses leakage:}
    Moving away from the origin by increasing $a$ and/or $|n-b|$ systematically reduces $|\alpha|$, which directly suppresses $|k_{pr}|$.
    \item \textbf{An optimal region exists for robust decoupling:}
    The chosen prototype parameters (marked point) lie in a plateau region where $|\alpha|$ is close to zero (e.g., $|\alpha|\approx 1.6^\circ$), leading to a near-zero $k_{pr}$ and $\eta_{\mathrm{roll}}$ close to unity.
\end{itemize}
Overall, optimizing $(a,b,n)$ to minimize $|\alpha|$ provides a direct structural route to suppressing the dominant leakage term $k_{pr}$ while maintaining high rolling effectiveness.

This analysis quantitatively proves that optimizing the mounting parameters $a$, $b$, and $n$ to minimize the contact angle $\alpha$ is the primary path to simultaneously reducing the leakage coefficient $k_{pr}$ and increasing the energy efficiency $\eta_{roll}$. This ``function-centric" spatial layout strategy provides the necessary structural foundation for achieving high-performance decoupling.

\subsection{Simulation-Based Validation of Coupling Effects on Dynamic Performance}

To validate the preceding theoretical analysis, we use inverse dynamics to compute the required motor command vector $\mathbf{\tau}_{\mathrm{cmd}}$ under different coupling strengths:
\begin{equation}
\mathbf{\tau}_{\mathrm{cmd}}=\mathbf{T}_{A}^{-1}\Big(\mathbf{M}(\mathbf{q})\,\ddot{\mathbf{q}}+\mathbf{C}(\mathbf{q},\dot{\mathbf{q}})\,\dot{\mathbf{q}}+\mathbf{G}(\mathbf{q})\Big).
\label{eq:C_t_i}
\end{equation}

We compare the performance of three actuation transmission matrices $\mathbf{T}_{A}$:
\begin{itemize}
    \item \textbf{Non-optimized baseline (zero separation):}
    $\mathbf{T}_{A}$ is computed from the geometry-derived contact angle corresponding to the baseline configuration ($a=b=n=0$), which yields a relatively large $|\alpha|$ (e.g., $|\alpha|\approx 21^\circ$) and thus a non-negligible $|k_{pr}|$ via \eqref{eq:k_r}. This baseline is fully specified by the structure rather than by arbitrary parameter choices.
    \item \textbf{Optimized design:}
    $\mathbf{T}_{A}$ is computed using the optimized structural parameters of the prototype, yielding $|\alpha|$ close to zero (e.g., $|\alpha|\approx 1.6^\circ$) and thus a near-zero $k_{pr}$.
    \item \textbf{Coupling stress test (parametric):}
    To illustrate how strong leakage amplifies control effort without claiming a specific prior robot, we keep the diagonal gains ($k_{pp},k_{rr}$) the same as the non-optimized baseline while imposing normalized leakage ratios
    $\kappa_{pr}=|k_{pr}|/k_{rr}$ and $\kappa_{rp}=|k_{rp}|/k_{pp}$ at a strong-but-reasonable level (e.g., $\kappa_{pr}=\kappa_{rp}=0.3$). The qualitative conclusions remain unchanged for $\kappa$ in the range 0.2--0.4.
\end{itemize}

\begin{figure}[thpb]
  \centering
  \includegraphics[width=0.48\textwidth]{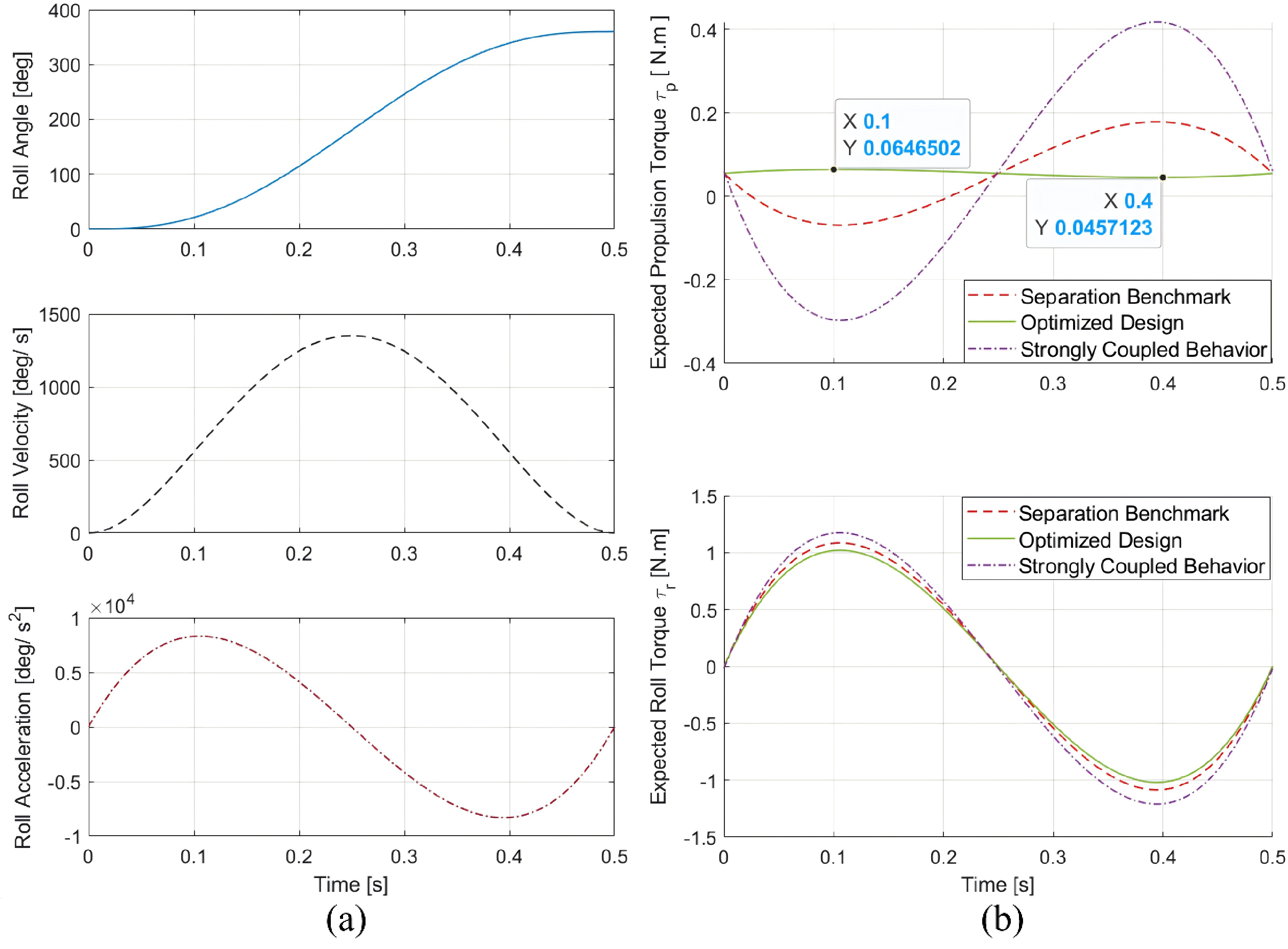}
  \caption{Inverse-dynamics simulation of coupling effects on required motor commands. (a) Prescribed high-dynamic roll trajectory with axial station-keeping against gravity. (b) Required propulsion command $\tau_p$ and roll command $\tau_r$ for three transmission configurations: geometry-derived non-optimized baseline (zero separation), optimized design, and a coupling stress test with imposed normalized leakage ratios.}
  \label{fig:Simu_D_A}
\end{figure}

The simulation task is a challenging, high-dynamic roll maneuver: the robot must maintain a constant axial position against gravity (g = 9.81 $\textup{m}/\textup{s}^2$) while executing a smooth, high-dynamic ``S-shaped" roll trajectory, as shown in Fig. \ref{fig:Simu_D_A}(a).

The simulation results in Fig.~\ref{fig:Simu_D_A}(b) reveal significant performance differences among the three configurations.
\begin{itemize}
    \item \textbf{Propulsion channel ($\tau_{p}$):}
    In this roll-dominant task with axial station-keeping against gravity, the ideal propulsion command $\tau_{p}$ should be a near-constant value primarily to counteract gravity. The \textbf{optimized design} yields an almost constant $\tau_p$, indicating negligible roll-induced leakage. The \textbf{non-optimized baseline (zero separation)} exhibits noticeable fluctuations correlated with roll acceleration, confirming that a non-negligible $k_{pr}$ increases compensation demand. The \textbf{coupling stress test} produces large, high-frequency fluctuations in $\tau_p$, illustrating how strong leakage can dramatically increase control effort and energy consumption.
    \item \textbf{Roll channel ($\tau_{r}$):}
    The \textbf{optimized design} and the \textbf{non-optimized baseline} yield similar $\tau_r$ profiles that track the prescribed roll acceleration, consistent with the fact that the main roll gain $k_{rr}\propto\cos\alpha$ changes only mildly for moderate $|\alpha|$. In contrast, the \textbf{coupling stress test} shows significant distortion in $\tau_r$ because any leakage from the propulsion channel (which continuously counters gravity) must be compensated by the roll motor.
\end{itemize}

The modeling and inverse-dynamics simulations support the main conclusion: optimizing structural parameters to minimize $|\alpha|$ suppresses the dominant leakage term $|k_{pr}|\propto|\sin\alpha|$ to a nearly negligible level, thereby reducing cross-channel compensation demand in dynamic maneuvers. The geometry-derived non-optimized baseline highlights the necessity of structural optimization even within a separated architecture, while the parametric coupling stress test illustrates how strong leakage can amplify required control effort. These results provide a coherent theoretical basis for the experimental validation in Section~V.

\section{EXPERIMENTAL VALIDATION}
Sections~III--IV establish a design-centric interpretation of decoupling: the mechanical layout determines the actuation transmission matrix $\mathbf{T}_{A}$, and the contact angle $\alpha$ governs the dominant roll-to-propulsion leakage term $k_{pr}\propto \sin\alpha$. Therefore, the experimental goal of this section is to validate two hardware-level claims. First, under a deliberately high-dynamic rolling disturbance in a vertical pipe, the propulsion channel should remain nearly invariant if $k_{pr}$ is sufficiently suppressed by a near-zero-$\alpha$ design. Second, the same decoupled behavior should translate into reliable traversal in a geometry- and material-diverse testbed where propulsion and posture adjustment must be executed concurrently through complex 3D fittings.

\begin{itemize}
    \item \textbf{Transmission-level verification of decoupling:} replicate the high-dynamic roll disturbance considered in Section~IV and measure whether the propulsion torque remains nearly invariant during rolling, using motor torque telemetry as the quantitative signature of roll-to-propulsion crosstalk.
    \item \textbf{System-level validation in non-ideal environments:} demonstrate repeatable, stall-free traversal on a multi-material testbed with complex 3D fittings (out-of-plane double elbow (OPDE) and vertical elbow), where posture correction and propulsion must occur concurrently.
\end{itemize}

\subsection{Experimental Testbed and System Architecture}
To assess the performance of the proposed decoupled unit under conditions approaching real-world industrial applications, a challenging experimental testbed was constructed. This section provides a detailed definition of the testbed's configuration and the basic experimental settings.

\subsubsection{Experimental Testbed}
\begin{figure}[t]
  \centering
  \includegraphics[width=0.35\textwidth]{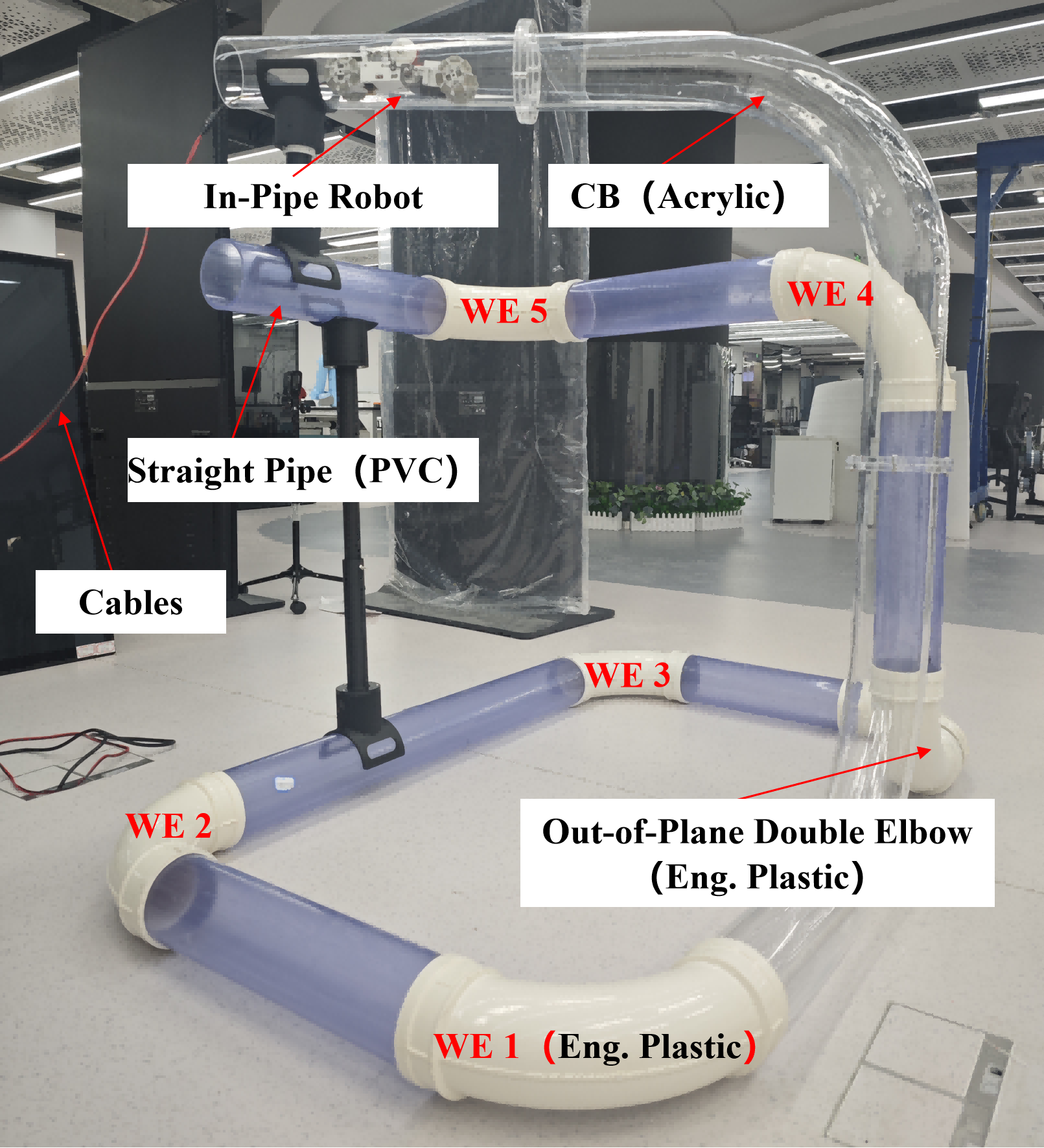}
  \caption{Platform for experiments in this work.}
  \label{fig:Test_Platform}
\end{figure}

The testbed has a total length of approximately 6 m with a uniform inner diameter of 100 mm. It is illustrated schematically in Fig. \ref{fig:Test_Platform}. The core design philosophy of the platform is to comprehensively test the robot's environmental adaptability, propulsion capability, and three-dimensional posture-adaptive ability by combining various materials and complex geometric fittings:
\begin{itemize}
\item \textbf{Material Diversity:}
To evaluate the robot's performance stability on different contact surfaces, the platform integrates three common materials found in industrial and laboratory settings: \textbf{Acrylic, PVC}, and \textbf{Engineering Plastic}. The introduction of these different materials aims to test the robustness of the robot's design with material variation as a key experimental variable.
\item \textbf{Geometric Complexity:}
The platform incorporates several geometric challenges, including \textbf{Continuous Bends (CB), White Elbows (WE), OPDE}, and \textbf{PVC Straight Pipes.}
\end{itemize}

The basic experimental procedure consisted of full round-trip traversals on the platform in Fig.~\ref{fig:Test_Platform} under teleoperation via the wireless handle (Fig.~\ref{fig:system-D}). Since teleoperated traversal time depends on operator strategy and proficiency (e.g., conservative posture correction versus aggressive speed pursuit), traversal time is treated as a secondary metric. To reduce sensitivity to outliers and operator-dependent variability, we summarize timing results using the median across repeated trials, while mission success rate is used as the primary indicator of passability and robustness. Fixed-view cameras recorded the entire process to extract segment-wise times in a consistent manner.

\subsubsection{Experimental System}
The complete experimental system consists of six main hardware components: the in-pipe robot unit, a regulated DC power supply, a custom umbilical cable, a laptop computer for control and data logging, a wireless operating handle, and a corresponding wireless receiver.

\subsubsection{System Architecture}
\begin{figure}[t]
  \centering
  \includegraphics[width=0.35\textwidth]{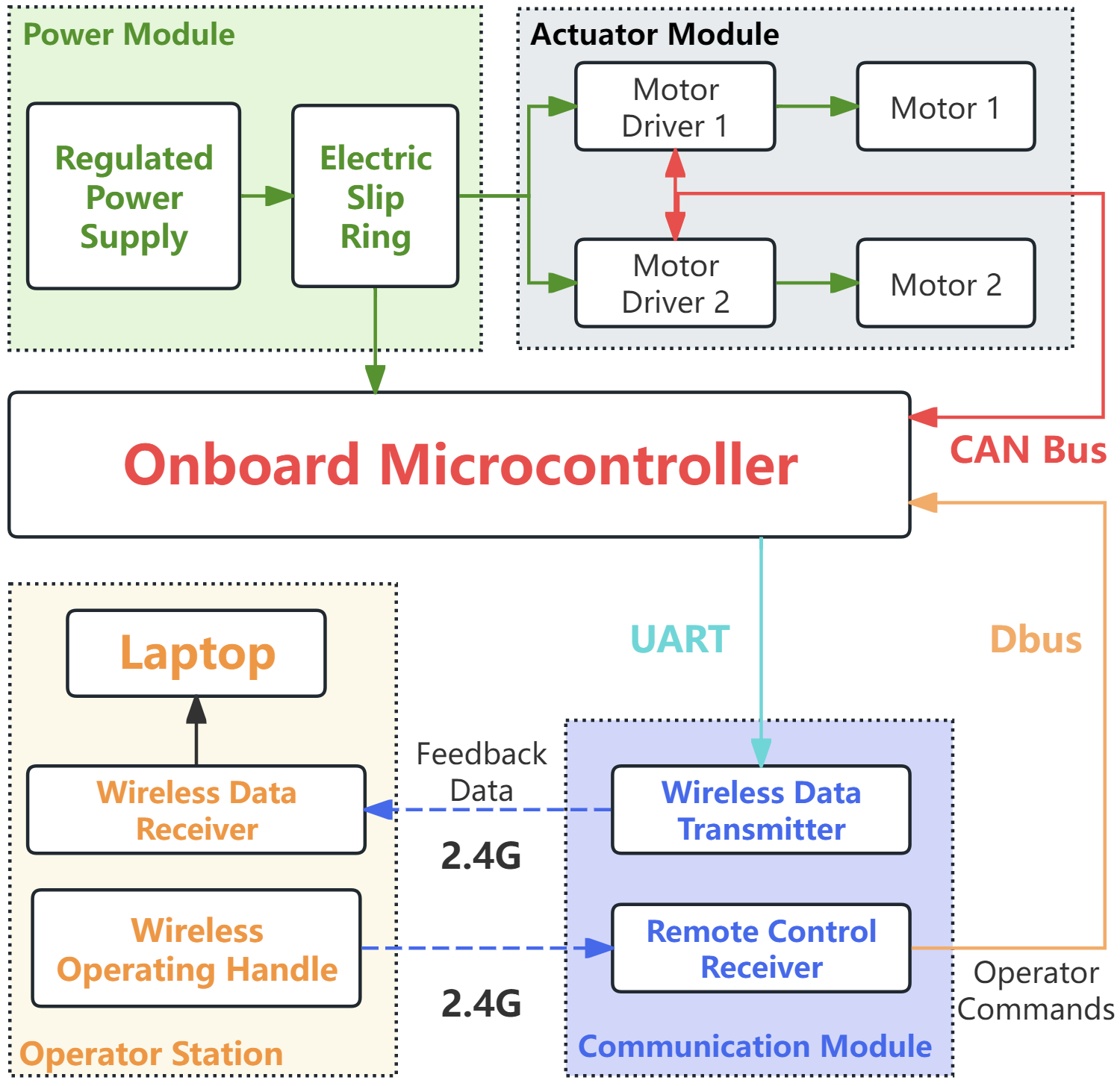}
  \caption{System architecture and data flow. Operator commands are transmitted wirelessly via a 2.4G link from the handle to the onboard receiver and processed by the microcontroller. Motor control is executed via a CAN Bus. System telemetry is sent back to a laptop for real-time monitoring via a separate wireless data link. The main power is supplied through a cable and an electric slip ring to allow for unrestricted robot rotation.}
  \label{fig:system-D}
\end{figure}

The system architecture, detailing the flow of power and data, is illustrated in Fig. \ref{fig:system-D}. The entire system is designed for direct, low-latency teleoperation, facilitating the rapid validation of the robot's mechanical capabilities. 

\subsection{Experimental Verification of Decoupling and Fault Tolerance}
The central implication of the $\alpha$-based transmission analysis is that, for an optimized near-zero $\alpha$ design, a high-dynamic roll maneuver should not inject a correlated disturbance into the propulsion channel (i.e., $k_{pr}$ should be negligible). To test this implication under a worst-case gravitational load, we conduct a vertical-pipe experiment in which the robot is held stationary while the rolling motor executes continuous revolutions. This experiment is designed to concurrently evaluate two core aspects: \textbf{dynamic crosstalk suppression} (the main claim of Section IV) and the \textbf{limit-state behavior} associated with the functional slip concept.

\begin{figure}[t]
  \centering
  \includegraphics[width=0.45\textwidth]{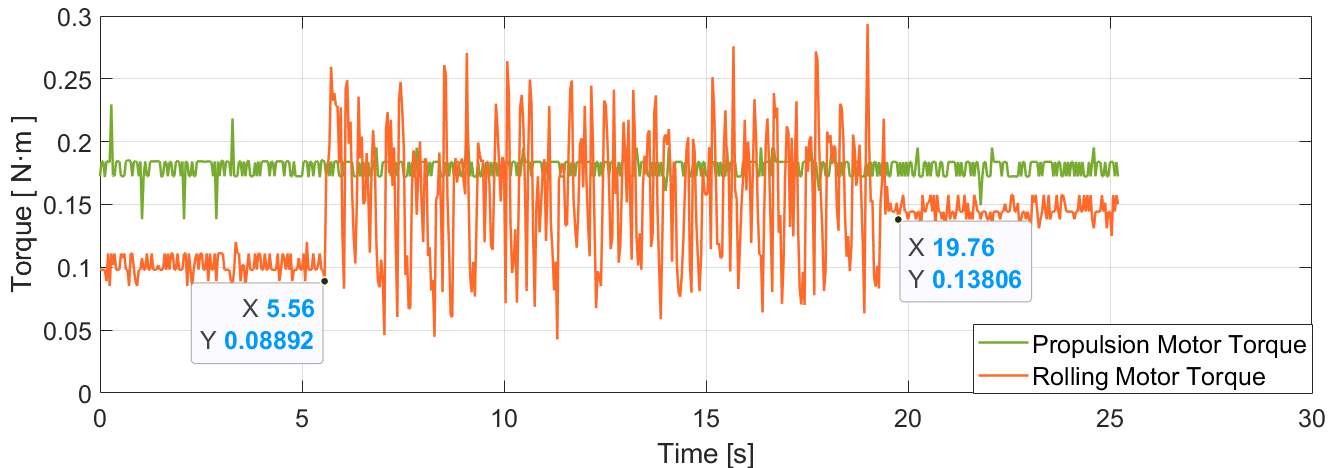}
  \caption{Experimental validation of kinematic decoupling under gravitational load. The robot was held stationary in a vertical pipe while the rolling motor was commanded to perform five continuous revolutions (from 5.56  $\textup{s}$ to 19.76 $\textup{s}$). 
  Despite the high and dynamic torque from the rolling motor (orange), the propulsion motor's torque (green) remained exceptionally stable with a standard deviation of less than 0.0077 $\textup{N}\cdot\textup{m}$, demonstrating minimal cross-talk between the two axes.}
  \label{fig:VT2}
\end{figure}

First, the torque telemetry in Fig.~\ref{fig:VT2} provides direct physical evidence of suppressed dynamic crosstalk. During the interval in which the roll motor executed five continuous revolutions (5.56~s to 19.76~s), the propulsion motor torque remained nearly invariant despite the highly dynamic roll torque. Quantitatively, the standard deviation of the propulsion motor torque over this rolling interval was below 0.0077~$\textup{N}\cdot\textup{m}$, indicating that roll excitation does not inject a correlated disturbance into the propulsion channel. This observation is consistent with the near-zero-$\alpha$ design point, for which the roll-to-propulsion leakage term $k_{pr}$ is expected to be small.

Notably, the measured baseline propulsion torque required for station-keeping in the vertical pipe is higher than the idealized prediction, which can be attributed to unmodeled effects such as drivetrain losses and additional friction sources. Importantly, the decoupling claim in this paper concerns the disturbance-induced variation of the propulsion torque during rolling rather than its absolute mean value. The exceptionally small torque variation under a high-dynamic roll disturbance is the key hardware signature supporting a small $k_{pr}$ at the physical transmission level.

\begin{figure}[t]
  \centering
  \includegraphics[width=0.45\textwidth]{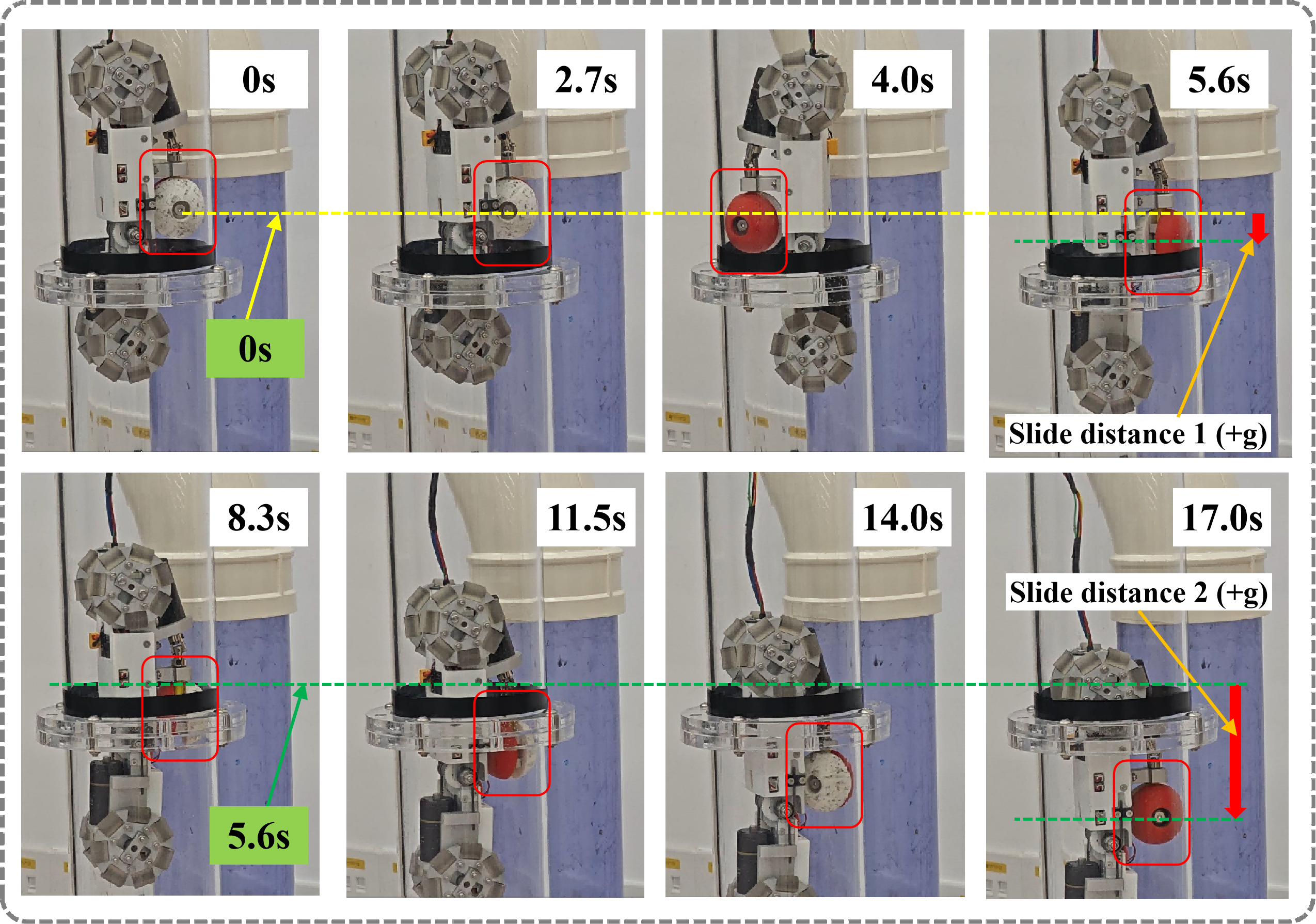}
  \caption{Kinematic sequence demonstrating the ``functional slip" mechanism during a vertical pipe test. While the robot performs continuous rolling (full sequence not shown), it exhibits a slow, controlled axial slip due to gravity. The slip distance was approximately 20 mm (slip distance 1) after the first revolution (5.6 $\textup{s}$) and accumulated to 95 $\textup{mm}$ (slip distance 2) after four revolutions (17.0 $\textup{s}$). This result validates our ``robustness-by-design" strategy, where a potential static failure (stalling) is transformed into a predictable, non-critical motion.}
  \label{fig: VT1}
\end{figure}

Second, Fig.~\ref{fig: VT1} visualizes a benign limit-state behavior associated with the functional-slip concept during rolling in a vertical pipe. While the static stability analysis in Section~III ensures feasibility under static-friction limits, a high-dynamic roll maneuver can drive the spherical-wheel contact toward kinetic friction, reducing the available axial friction compared to the static case. As a result, gravity cannot be fully balanced and a slow, predictable axial slip occurs while rolling continues. In the experiment, the slip distance was approximately 20~mm after the first revolution and accumulated to about 95~mm after four revolutions. From a system-design perspective, this behavior supports a fault-tolerant interpretation: instead of an abrupt stall or stick--slip instability near challenging configurations, the system exhibits a low-energy, non-catastrophic motion that preserves roll-channel operation and yields a predictable outcome.

In summary, the vertical-pipe experiment validates the two key physical-layer claims required by the new narrative: (i) under high-dynamic roll disturbances, the propulsion channel remains effectively isolated (quantitatively evidenced by the low torque variation), and (ii) the functional-slip behavior emerges as a benign limit-state that does not compromise the roll maneuver. Together, these results provide hardware-level support for the $\alpha$-guided decoupling framework.

\subsection{Holistic Performance Evaluation in Complex Environments}
After validating decoupling at the actuation-transmission level, we evaluate whether the same design principles remain effective in complex environments that combine material variation and 3D fittings. In such scenarios, posture adjustment and propulsion must occur concurrently; hence, successful traversal with continuous motion provides an operational demonstration of the decoupled architecture.

\begin{figure}[t]
  \centering
  \includegraphics[width=0.45\textwidth]{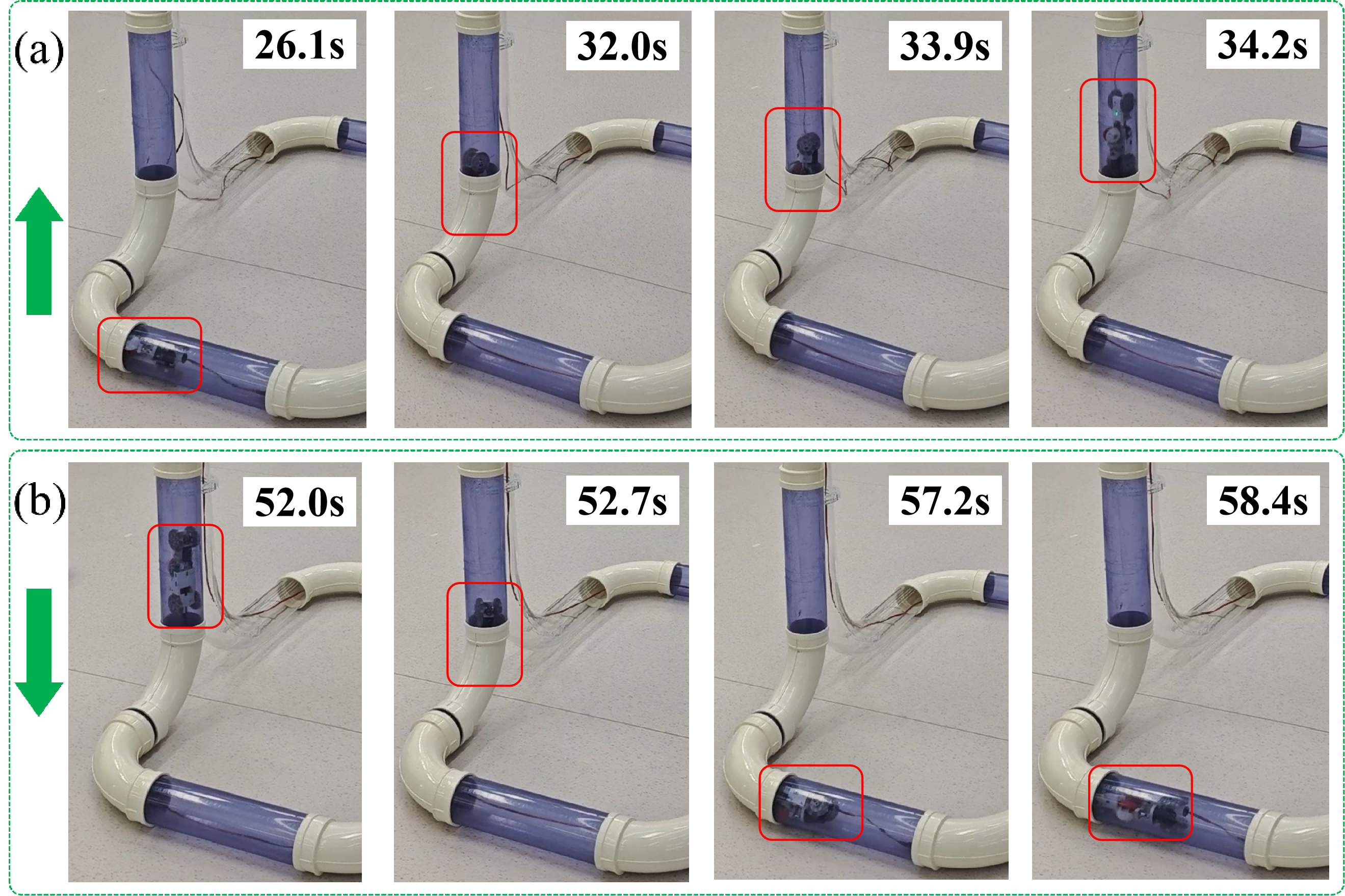}
  \caption{Kinematic sequence of the robot traversing the OPDE. (a) Ascending pass. (b) Descending pass. The continuous motion and change in roll angle demonstrate the concurrent execution of propulsion and rolling.}
  \label{fig:OPDE}
\end{figure}

\begin{figure}[t]
  \centering
  \includegraphics[width=0.45\textwidth]{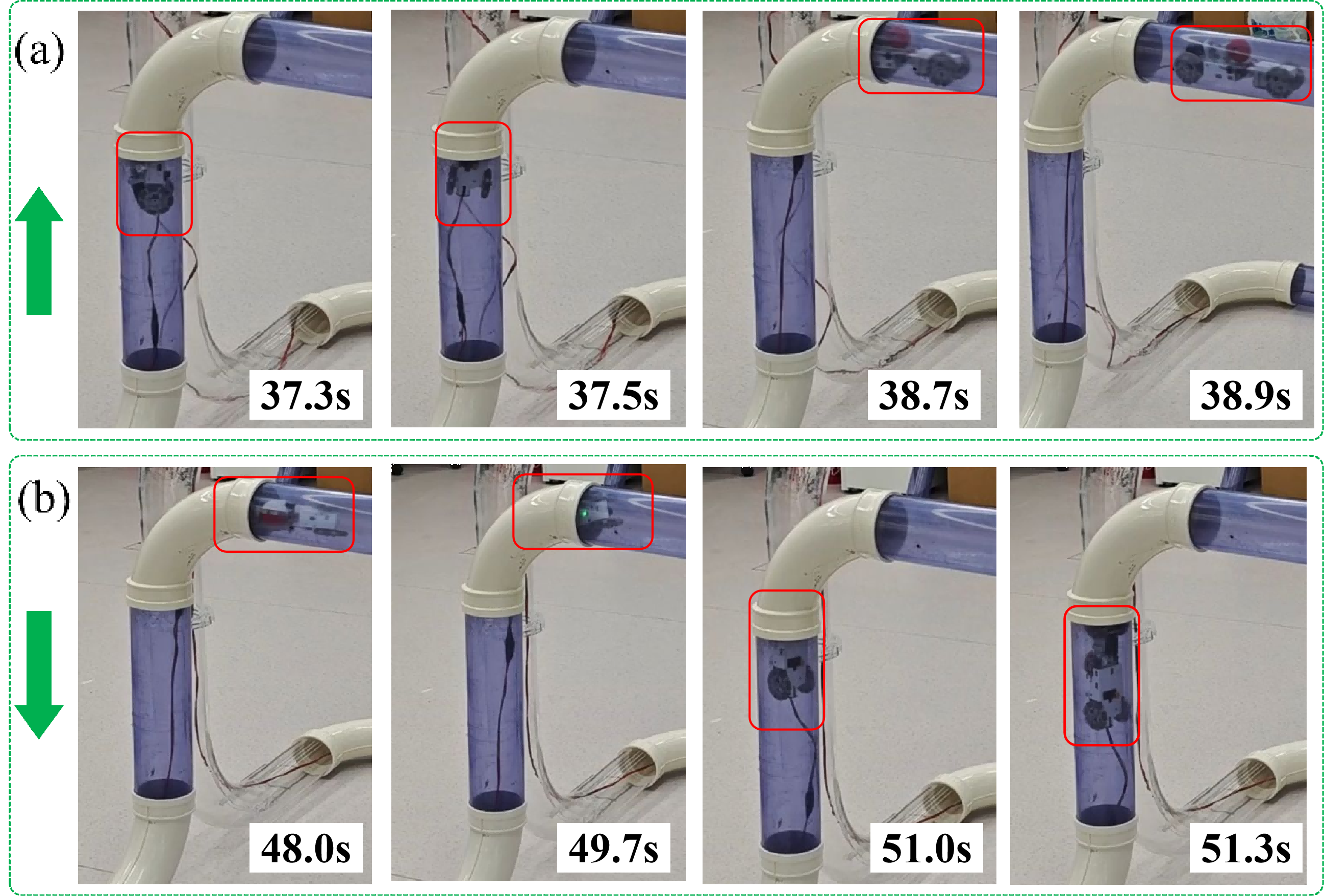}
  \caption{Demonstration of posture-adaptive correction in the vertical elbow (WE4). (a) Ascending pass. (b) Descending pass. The robot enters with a visible misalignment but rapidly self-corrects its posture within 0.2 s, showcasing the dynamic response of the decoupled rolling system.}
  \label{fig:Straight}
\end{figure}

The robot was tested on the multi-material platform defined in Fig. \ref{fig:Test_Platform}. Figures \ref{fig:OPDE} and \ref{fig:Straight} highlight two representative challenges: the OPDE traversal and a vertical elbow (WE4). These cases emphasize different aspects of the system: OPDE requires continuous posture adaptation during propulsion in a 3D geometry, while the vertical elbow tests rapid posture correction against gravity.

In the OPDE traversal (Fig.~\ref{fig:OPDE}), the robot exhibited continuous, stall-free motion while propulsion and rolling were executed concurrently, demonstrating posture-adaptive agility in a complex 3D fitting. In the vertical elbow test (Fig.~\ref{fig:Straight}), the robot entered with a severe posture misalignment but rapidly self-corrected within 0.2~s against gravity, highlighting the high-response characteristics of the decoupled roll system. Because the platform is teleoperated, traversal time is influenced by the operator’s strategy (e.g., prioritizing conservative posture correction versus speed). Therefore, the primary system-level outcome reported here is consistent passability without stalling across repeated trials, while timing statistics are reported as secondary descriptors.

For a compact quantitative summary, the robot's performance was assessed via representative round-trip traversals, and key metrics are reported in Table~\ref{tab:final_gravity_bidirectional}. Crucially, a 100\% mission success rate was achieved over ten independent trials, providing initial statistical support for the design's reliability.

Because the platform trials are teleoperated, traversal time is operator-dependent and is treated as a secondary descriptor; the primary system-level outcome reported here is mission passability (10/10 successful round trips).

\begin{table*}[t!]
\centering
\begin{threeparttable}
\caption{Performance Analysis of Bidirectional Traversal Under Varying Gravitational Effects (Segment-wise time/speed from one randomly selected trial among 10 teleoperated round-trip trials)}
\label{tab:final_gravity_bidirectional}
\begin{tabularx}{15cm}{c l c c c >{\centering\arraybackslash}X}
\toprule
\textbf{Pass Direction} & \textbf{Traversal Section} &
\textbf{Gravitational Effect} & \textbf{Length ($\textup{m}$)} & \textbf{Time ($\textup{s}$)} & \textbf{Avg. Speed ($\textup{m}/\textup{s}$)} \\
\midrule
\multirow{5}{*}{\textit{\textbf{Forward Pass}}} 
& 1.  CB (\textbf{Acrylic}) & {Assist ($+\textup{g}$)} & 2.60 & 12.0 & 0.22 \\
& 2. WE 1, 2, 3 (Eng. Plastic) & Horizontal ($\sim$ $0\textup{g}$) & 0.47 (avg) & 2.1 (avg) & 0.22 \\
& 3. \textbf{OPDE} (Eng. Plastic)
&{Inhibit ($-\textup{g}$)} & \textbf{0.94} & \textbf{8.1} & \textbf{0.12} \\
& 4. WE 4 (Eng. Plastic) & {Inhibit ($-\textup{g}$)} & 0.47 & 1.6 & 0.29 \\
& 5. WE 5 (Eng. Plastic) & Horizontal ($\sim$ 0$\textup{g}$) & 0.47 & 1.8 & 0.26 \\
\midrule

\multirow{5}{*}{\textit{\textbf{Backward Pass}}}
& 6. WE 5  (Eng. Plastic)& Horizontal ($\sim$ 0$\textup{g}$) & 0.47 & 1.7 & 0.28 \\
& 7. WE 4  (Eng. Plastic)& {Assist ($+\textup{g}$)} & 0.47 & 3.3 & 0.14 \\
& 8. \textbf{OPDE} (Eng. Plastic)  &{Assist ($+\textup{g}$)} & \textbf{0.94} & \textbf{6.4} & \textbf{0.15} \\
& 9. WE 3, 2, 1 (Eng. Plastic) & Horizontal ($\sim$ $0\textup{g}$) & 0.47 (avg) & 2.2 (avg) & 0.21 \\
& 10. CB (\textbf{Acrylic}) & {Inhibit ($-\textup{g}$)} & 2.60 & 4.6 & 0.57 \\
\bottomrule
\end{tabularx}
\begin{tablenotes}[flushleft, online] 
\footnotesize

\item[\textbf{Note:}] The section-wise time and speed reported in Table~\ref{tab:final_gravity_bidirectional} are computed from one randomly selected round-trip trial out of 10 independent teleoperated trials (the trial was selected without regard to performance). Because teleoperated traversal time depends on operator strategy and proficiency, these time/speed values are provided as an illustrative segment-level description rather than a primary metric. Mission-level passability was evaluated over all 10 trials, and a 100\% success rate (10/10) was achieved, \emph{i.e.}, each trial completed the full round trip without external assistance or manual intervention. Videos of the trials are included in the supplementary material. The ``Gravitational Effect'' column indicates whether gravity assisted ($+\textup{g}$, downhill), inhibited ($-\textup{g}$, uphill), or had a neutral effect on propulsion.
\end{tablenotes}
\end{threeparttable}
\end{table*}

The data reveals the robot's predictable dynamic performance under varying gravitational loads. As expected, traversal speed decreased during ascent (e.g., OPDE, $-\textup{g}$) and increased during descent ($+\textup{g}$), demonstrating stable control.

More interestingly, the WE4 segment highlights a posture-dominated timing characteristic: the robot may spend additional time on deliberate self-alignment even when gravity assists propulsion. Such behavior is consistent with a “posture-first” operational strategy in which maintaining a correct posture is prioritized over maximizing speed. Since the trials were teleoperated, traversal time also depends on operator proficiency and strategy; nevertheless, the consistent completion of the OPDE and vertical elbow traversals without stalling across repeated trials indicates that passability is maintained even under conservative (slower) operation.

In summary, the experiments support the paper’s $\alpha$-centered decoupling claim in a manner relevant to automation. The vertical-pipe disturbance test provides a direct transmission-level signature: rolling can be excited dynamically while the propulsion torque remains nearly invariant, indicating negligible roll-to-propulsion leakage consistent with a near-zero $\alpha$ design. The multi-material platform trials then demonstrate that this reduced crosstalk translates into repeatable system-level behavior in complex 3D fittings, where posture correction and propulsion must be executed concurrently. Together, these results show that the proposed structural optimization does not merely improve traversal qualitatively; it yields a measurable reduction in actuation interference that simplifies the control burden and increases operational robustness in realistic pipe geometries.

\section{Conclusion}
This paper presented a V-shaped in-pipe robot enabled by a joint-axis-and-wheel-separation architecture to physically decouple axial propulsion and rolling reorientation. A unified geometric--static--dynamic framework was developed to make the structure--performance link explicit, where the contact angle $\alpha$ serves as the key geometric variable governing the dominant roll-to-propulsion leakage term in the actuation transmission matrix and the roll-channel energy efficiency. The static model provided a stability-domain guideline under frictional uncertainty, establishing a reliable physical operating region for the decoupled actuation.

Experimental results supported the proposed $\alpha$-centered narrative. Under a high-dynamic roll disturbance in a vertical pipe, the propulsion torque remained nearly invariant, indicating negligible dynamic crosstalk consistent with a near-zero-$\alpha$ design. Traversal tests on a multi-material platform with complex 3D fittings further demonstrated repeatable system-level operation with concurrent propulsion and posture correction. Future work will extend the framework to explicitly account for curvature-dependent contact conditions and to integrate sensing and autonomy for long-range inspection in unstructured pipelines.

\appendices
\section{CLOSED-FORM SOLUTION OF THE STATIC NORMAL FORCES}
\label{app:static_closed_form}

In \eqref{eq: eq2}--\eqref{eq: eq3}, the normal-force moments can be written as linear functions of the unknown normal forces:
\begin{equation}
M_{N0}= \ell_{N0}\,F_{N0},\quad
M_{N1}= \ell_{N1}\,F_{N1},\quad
M_{N2}= \ell_{N2}\,F_{N2}
\end{equation}
where each $\ell_{Ni}$ is the signed moment arm of the corresponding normal force about the joint $O$ (positive counter-clockwise), computed by the vector moment definition:
\begin{equation}
\ell_{Ni} = \left(\mathbf{r}_{Ni}\times \mathbf{n}_{i}\right)\cdot \mathbf{e}_z ,
\end{equation}
with $\mathbf{r}_{Ni}$ being the position vector from $O$ to the application point of $F_{Ni}$ in Fig.~\ref{fig:AlphaDef}, $\mathbf{n}_i$ the inward unit normal direction at that contact, and $\mathbf{e}_z$ the out-of-plane unit vector.

Similarly, each gravitational moment is written as:
\begin{equation}
M_{Gi}=\ell_{Gi}\,G_i,\qquad
\ell_{Gi}=\left(\mathbf{r}_{Gi}\times \mathbf{g}\right)\cdot \mathbf{e}_z ,
\end{equation}
where $\mathbf{r}_{Gi}$ is the position vector from $O$ to the center of mass of component $i$, and $\mathbf{g}$ is the gravity direction (along the pipe axis for the vertical-pipe case).

With the above definitions, \eqref{eq: eq3} yields $F_{N2}$ directly:
\begin{equation}
F_{N2} = -\frac{M_{G2}+M_{G4}+M_J}{\ell_{N2}} .
\label{eq:app_FN2}
\end{equation}
Substituting \eqref{eq:app_FN2} into \eqref{eq: eq1} gives $F_{N0}=F_{N1}+F_{N2}$. Plugging this relation into \eqref{eq: eq2} yields:
\begin{equation}
(\ell_{N0}+\ell_{N1})F_{N1} + \ell_{N0}F_{N2} + (M_{G0}+M_{G1}+M_{G3})-M_J = 0 .
\end{equation}
Therefore,
\begin{equation}
F_{N1}=
\frac{M_J-(M_{G0}+M_{G1}+M_{G3})-\ell_{N0}F_{N2}}
{\ell_{N0}+\ell_{N1}},
\label{eq:app_FN1}
\end{equation}
and finally,
\begin{equation}
F_{N0}=F_{N1}+F_{N2}.
\label{eq:app_FN0}
\end{equation}

Equations \eqref{eq:app_FN2}--\eqref{eq:app_FN0} constitute the closed-form solution of the static normal forces used in Section III.

\section{DEFINITION OF STABILITY SAFETY MARGIN AND TRACTION RESERVE}
\label{app:safety_margin}

For the vertical-pipe case in Section III, the maximum total static friction force that can be generated is:
\begin{equation}
f_{max,total}=\mu_sF_{N0}+\mu_o(F_{N1}+F_{N2}).
\end{equation}
Define the stability safety margin as:
\begin{equation}
S = \frac{f_{max,total}}{G_{total}}.
\end{equation}
The critical stability boundary in Fig.~\ref{fig: 3D-SP-K20} corresponds to $S=1$.

The ``traction reserve against gravity'' reported in Section III is defined as the remaining fraction of available friction after balancing gravity:
\begin{equation}
\text{Traction reserve} = 1-\frac{G_{total}}{f_{max,total}} = 1-\frac{1}{S}.
\end{equation}
For the nominal design point with $S=2.3$, this gives $1-1/2.3\approx 0.565$, i.e., a 56.5\% traction reserve.

\section{INERTIAL DECOUPLING OF THE IDEAL V-SHAPED STRUCTURE}
\label{app:inertial_crosstalk}

This appendix provides the rigid-body-level proof that the V-shaped structure eliminates translation--rotation inertial crosstalk in the ideal model. Using generalized coordinates $\mathbf{q}=[z(t),\phi(t)]^{T}$, the kinetic energy of each link consists of translational and rotational parts. By rewriting the total kinetic energy in the quadratic form of generalized velocity $\dot{\mathbf{q}}$, the inertia matrix can be obtained as:

\begin{equation}
\begin{aligned}
\mathbf{M} & = \left[\begin{array}{cc}
M_{z}^z         & M_{z}^{\phi}  \\
M_{\phi}^{z}    &  M_{\phi}^{\phi} 
\end{array} \right] \\
& =\left[\begin{array}{cc}
m_1+m_2         & 0  \\
0               & \sum_{i=1}^{2} m_{i}\left(\frac{L_{i}^{2}}{4} \sin ^{2} \theta_{i}+\frac{L_{i}^{2}}{12}\right)
\end{array}\right]
\end{aligned}
\label{eq: V_C}
\end{equation}

The off-diagonal terms $M^{\phi}_{z}$ and $M^{z}_{\phi}$ represent inertial coupling between the translational acceleration ($\ddot{z}$) and the rolling torque ($M_{r}$), and between the rolling acceleration ($\ddot{\phi}$) and the propulsive force ($M_{p}$). The result $M^{\phi}_{z}=M^{z}_{\phi}=0$ proves that the V-shaped structural design fundamentally eliminates these two types of inertial crosstalk at the ideal rigid-body level. Therefore, the dominant coupling analyzed in Section IV is attributed to the physical actuation transmission process captured by $\mathbf{T_A}$.

\bibliography{m_bib}

\bibliographystyle{IEEEtran}

\end{document}